\documentclass[11pt]{article}

\usepackage[final]{acl}

\usepackage{times}
\usepackage{latexsym}

\usepackage[T1]{fontenc}

\usepackage[utf8]{inputenc}

\usepackage{microtype}

\usepackage{inconsolata}

\usepackage{graphicx}

\usepackage{booktabs} 
\usepackage{multirow} 

\usepackage{geometry}
\usepackage{xcolor}
\usepackage{tcolorbox}
\usepackage{caption}
\usepackage{amsmath}
\usepackage{bbm}
\usepackage{amsfonts}

\definecolor{promptbg}{RGB}{247, 249, 252} 
\definecolor{promptframe}{RGB}{66, 133, 244} 

\title{\textbf{SAGE}: Semantic Anchor-Guided Evolution for Grounded Medical QA Data Synthesis}

\author{
  \textbf{Chuan Li\textsuperscript{1}},
  \textbf{Chengyu Wang\textsuperscript{2}},
  \textbf{Cen Chen\textsuperscript{1}\thanks{Corresponding author}}, 
  \textbf{Ye Lyu\textsuperscript{1}},
  \textbf{Mingyuan Fan\textsuperscript{1}},
  \textbf{Ming Gao\textsuperscript{1}} \\
  \textsuperscript{1}East China Normal University, Shanghai, China \\
  \textsuperscript{2}Alibaba Group, Hangzhou, China \\
  \texttt{\{lichuan, yelyu, mingyuan\_fmy\}@stu.ecnu.edu.cn} \\
  \texttt{chengyu.wcy@alibaba-inc.com, \{cenchen,mgao\}@dase.ecnu.edu.cn}
}

\begin{document}
\maketitle
\begin{abstract}
Developing reliable models for clinical tasks, such as Medical Question Answering (QA), is severely constrained by the limited availability of high-quality, expert-annotated training data. This challenge is exacerbated by stringent privacy requirements and the impracticality of utilizing large open-source corpora or proprietary cloud APIs within resource-limited clinical settings.
To address these obstacles, we introduce SAGE (\textit{Semantic Anchor-Guided Evolution}), a novel data synthesis framework that enables small, locally deployed models to generate high-quality medical training data. SAGE leverages lightweight, publicly available taxonomies such as MeSH as semantic anchors, imposing a structured prior to effectively guide and ground the data generation process. At its core, SAGE iteratively interleaves atomic (individual concept-based) and associative (relation-based) synthesis, bootstrapping training data from minimal seeds. This approach eliminates the need for large collections of medical documents or reliance on external APIs, providing a practical solution for on-premises data creation.
Extensive experiments across multiple medical question-answering benchmarks demonstrate that models fine-tuned with SAGE-synthesized data consistently outperform those trained using self-derived or conventional document-based paradigms, highlighting tangible improvements in data efficiency and resource utilization for medical LLM development. Code is available at \url{https://github.com/DIaacKr/SAGE}.
\end{abstract}

\section{Introduction}

Large language models (LLMs) have demonstrated remarkable capabilities, prompting significant interest in their deployment within high-stakes, knowledge-intensive domains such as medicine~\cite{zhang2024ultramedical,guo2025deepseek,qwenQwen25TechnicalReport2025,yang2025qwen3,xu2025lingshu,huangM1UnleashPotential2025,yu2025finemedlmo,wang2025baichuan,chen2025towards}. However, further progress is hampered by the scarcity of high-quality, expert-annotated datasets. In medicine, assembling large-scale training corpora is not only expensive and time-consuming, but also subject to stringent privacy and ethical constraints~\cite{price2019privacy,kaissis2020secure,rajpurkar2022ai,thirunavukarasu2023large}. This poses a fundamental bottleneck for transforming general-purpose LLMs into reliable medical experts.

Synthetic data generation has emerged as a promising direction; yet, existing approaches predominantly rely on proprietary, large-scale LLMs as generative engines~\cite{wang2023self,gudibande2023false}. Such dependence introduces challenges regarding data privacy, operational costs, and practical feasibility in resource-constrained environments, including hospitals~\cite{carlini2021extracting}. Addressing this underexplored issue requires a practical and resource-efficient method for synthesizing high-quality training data using strictly limited, local seed resources, thereby eliminating dependency on external APIs or massive proprietary corpora~\cite{labrak2024biomistral}.

\begin{figure}
    \centering
    \includegraphics[width=0.98\linewidth]{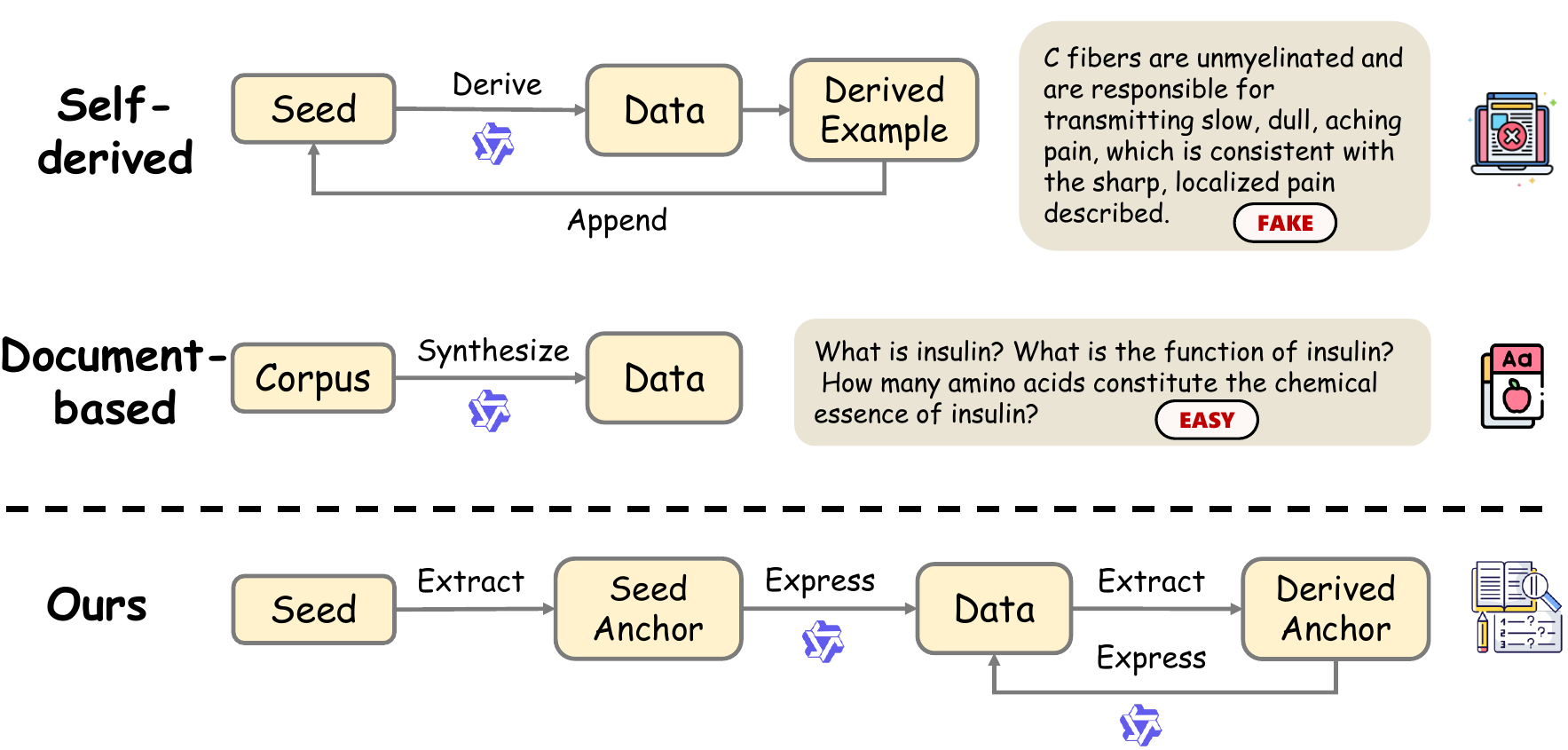}
    \caption{Comparison of data synthesis paradigms. Conventional methods either derive data using implicit parametric knowledge or synthesize data from document corpus. In contrast, SAGE synthesizes high-quality medical data via semantic anchor.}
    \vspace{-.5em}
    \label{fig:intro}
\end{figure}

Prior work in data synthesis primarily falls into two paradigms, each with clear limitations, as shown in Figure~\ref{fig:intro}. The \emph{self-derived} paradigm leverages implicit parametric knowledge to efficiently generate diverse data~\cite{wang2023self,xu2024wizardlm,yu2025cot}; yet, the absence of external grounding can result in unanchored outputs and hallucinations, which is an unacceptable risk for clinical applications. Conversely, the \emph{document-based} paradigm grounds generation in authoritative texts to promote factual accuracy~\cite{maini2024rephrasing,miao2025easy}. However, document-based methods require extensive preprocessing of large corpora, which are often unavailable or unstructured in clinical contexts. Furthermore, document-based approaches tend to encourage mere ``memorization'' of text, rather than the ``integration'' of complex clinical reasoning, constraining model expressivity and applicability.



We contend that structured medical taxonomies, Medical Subject Headings (MeSH)\footnote{\url{https://www.ncbi.nlm.nih.gov/mesh/}}, represent a widely accessible yet underutilized resource. Unlike lengthy documents or corpora, MeSH serves as a standardized dictionary, which is a topological scaffold of medical concepts, that is both lightweight and authoritative. In this work, we introduce SAGE (\emph{Semantic Anchor-Guided Evolution}), a grounded QA data synthesis framework specifically designed for resource-constrained environments. SAGE treats MeSH concepts as \emph{semantic anchors} to guide and ground the generative process. This neuro-symbolic approach enables small, locally deployed models to produce logically rigorous content, adhering to a structured prior while reducing both computational and data overhead.

SAGE orchestrates data synthesis leveraging semantic anchors through a dual-track, parallel, iterative process that balances exploitation and exploration, as shown in Figure~\ref{fig:intro}. To ensure factual precision, SAGE employs \emph{atomic generation}, targeting individual semantic anchors via ontology definitions to produce definition-centric questions and mitigate hallucination risk. Meanwhile, SAGE utilizes \emph{associative generation}, interweaving multiple concepts as anchor to construct coherent medical scenarios requiring multi-step reasoning. Crucially, these tracks operate concurrently, maximizing the coverage of the concept space of core concepts; they are refined through ongoing iterative calibration that broadens the semantic scope of the synthesized dataset. This evolutionary loop enables bootstrapping from a minimal seed set, organically expanding from atomic facts to complex associative reasoning. As a result, SAGE cultivates comprehensive and valuable training corpora from compact, structured sources.



Our method achieves state-of-the-art performance across a broad range of medical QA benchmarks, with an average accuracy of 68.12\% on five widely used medical QA datasets, outperforming other self-derived and document-based methods. Notably, it also exceeds the results achieved by several high-quality open-source datasets and models trained on massive data.

In summary, our main contributions are:
\begin{itemize}
    \item We introduce SAGE, a Semantic Anchor-Guided Evolution framework for synthesizing high-quality, grounded medical QA data using small models and public taxonomies, without reliance on large corpora or external APIs.
    \item We propose a parallel, iterative synthesis strategy that integrates atomic generation for factual coverage and associative generation for reasoning complexity, dynamically balancing exploitation and exploration.
    \item SAGE-grounded data substantially improve factual reliability and consistency. Models fine-tuned on SAGE data consistently outperform strong baselines on medical QA benchmarks in resource-constrained settings.
\end{itemize}

\section{Related Work}

\noindent\textbf{Self-Derived Data Synthesis.}  
Self-bootstrapping data synthesis methods, which autonomously generate large-scale training data, have attracted increasing research interest as alternatives to manual annotation~\cite{li2023two,yu2023metamath,luo2023wizardmath,xu2024wizardlm,tang2024mathscale,li2024common,zeng2024automatic,huang2025key,liu2025augmenting,riaz2025metasynth,zhao2025promptcot,zhao2025promptcot2}. A representative example is \textit{Self-Instruct}~\cite{wang2023self}, which prompts LLMs to produce diverse instructions and corresponding responses, expanding training data from a small seed set. Subsequent variants, including \textit{Chain-of-Thought Self-Instruct}~\cite{yu2025cot}, introduce ``think-then-generate'' pipelines with automatic filtering for higher-quality examples. However, these approaches lack rigorous mechanisms for factual control: filtering criteria typically prioritize semantic richness or complexity rather than factuality. In high-stakes fields such as medicine, this limitation becomes critical, leading to severe factual hallucinations and adversely affecting model fine-tuning.
 
\noindent\textbf{Document-Based Data Synthesis.}  
Document-based approaches utilize structured or unstructured corpora to synthesize training data for LLMs~\cite{zhang2023self,yehudai2024genie,koksal2024longform,maini2024rephrasing,yang2024synthetic,dedhia2025bottom,chen2025graphgen}. For instance, \textit{Easy Datasets}~\cite{miao2025easy} guides models to create question-answer pairs from the same source text by assigning varied ``genre-audience'' roles, thereby enhancing the diversity and scale of synthetic data for domain adaptation. Similarly, \textit{Wrap}~\cite{maini2024rephrasing} leverages off-the-shelf LLMs to rephrase noisy web text into higher-quality synthetic formats, improving both compute efficiency and downstream performance. While these methods ensure factual grounding through anchoring to source documents, their success heavily relies on access to large, high-quality document collections. In low-resource scenarios where such corpora are unavailable, the practical utility of document-based synthesis is limited.

\section{Methodology}
\label{sec:method}

We present \textbf{Semantic Anchor-Guided Evolution} (\textbf{SAGE}), a medical QA synthesis framework that alternates between neural generation in text space and symbolic projection onto a fixed medical taxonomy. Unlike unconstrained self-generation, SAGE uses MeSH concepts as semantic anchors to organize the generation trajectory while retaining the flexibility of a language model to express relations that are not explicitly encoded as taxonomy edges.

\subsection{Problem Formulation and Neural-Symbolic Evolution}
\label{sec:formulation}




Our objective is to synthesize a medical question-answering dataset, denoted by $\mathcal{D}_{\mathrm{syn}}$, using a locally deployed language model $\mathcal{M}$ and a small seed dataset $\mathcal{D}_{\mathrm{seed}}$. We use an authoritative medical taxonomy, such as MeSH, as an external structured prior. Let
\begin{equation}
\mathcal{T} = (\mathcal{V}, \mathcal{E})
\end{equation}
denote the fixed taxonomy, where $\mathcal{V}$ is the set of medical concept nodes and $\mathcal{E}$ represents their hierarchical relations.

SAGE maintains an evolving population of active anchor configurations
\begin{equation}
\mathcal{S}^{(t)} \subseteq 2^{\mathcal{V}} \setminus \{\emptyset\}
\end{equation}
at epoch $t$. Each anchor configuration $\mathbf{A} \in \mathcal{S}^{(t)}$ contains one or more MeSH concepts and specifies the semantic scope of a generation step. 

Given an anchor configuration $\mathbf{A}$, the language model generates a candidate QA instance $x=(q_x,r_x,y_x)$, where $q_x$ is the question, $r_x$ is the generated reasoning, and $y_x$ is the answer label. After candidate selection, $q_x$ is mapped back to the taxonomy and may contribute an updated anchor configuration to the next epoch.

This process can be viewed as neural-symbolic evolution over a fixed ontological scaffold. The language model serves as a neural variation operator that explores possible medical expressions and relations, whereas the taxonomy provides a discrete concept space for anchoring, projection, and population update. SAGE does not perform a hand-designed traversal of taxonomy edges; instead, semantic exploration is carried out by the language model in text space and is iteratively consolidated in the MeSH concept space. Accordingly, the taxonomy itself remains fixed, while the active population $\mathcal{S}^{(t)}$ evolves across epochs.




\subsection{MeSH-Structured Semantic Anchors}
\label{sec:semantic_anchor}

We define a taxonomy-aware mapping function
\begin{equation}
\phi_{\mathcal{T}} : \mathcal{X}_{q} \rightarrow 2^{\mathcal{V}},
\end{equation}
which maps a medical question $q \in \mathcal{X}_{q}$ to a set of MeSH concepts.
In implementation, $\phi_{\mathcal{T}}$ performs synonym-expanded terminology matching and retains concepts according to predefined MeSH category and hierarchy-depth constraints.
Additional details and evaluations of the mapping function are provided in Appendix~\ref{ap:mapping_evaluation}.

The set $\phi_{\mathcal T}(q_x)$ is referred to as the semantic anchor configuration of instance $x$. These anchors do not prescribe a complete reasoning path. Instead, they provide a structured semantic scaffold that specifies which medical concepts should organize the generation. This allows the model to draw on its parametric knowledge while keeping the iterative synthesis trajectory connected to a curated medical concept space.

The initial anchor population is constructed as
\begin{equation}
\mathcal{S}^{(0)}
=
\left\{
\phi_{\mathcal{T}}(q_x)
\mid
x \in \mathcal{D}_{\mathrm{seed}},
\ \phi_{\mathcal{T}}(q_x) \neq \emptyset
\right\}.
\end{equation}

\subsection{Dual-Track Neural Generation}
\label{sec:dual_track}

\begin{figure*}
    \centering
    \includegraphics[width=0.85\linewidth]{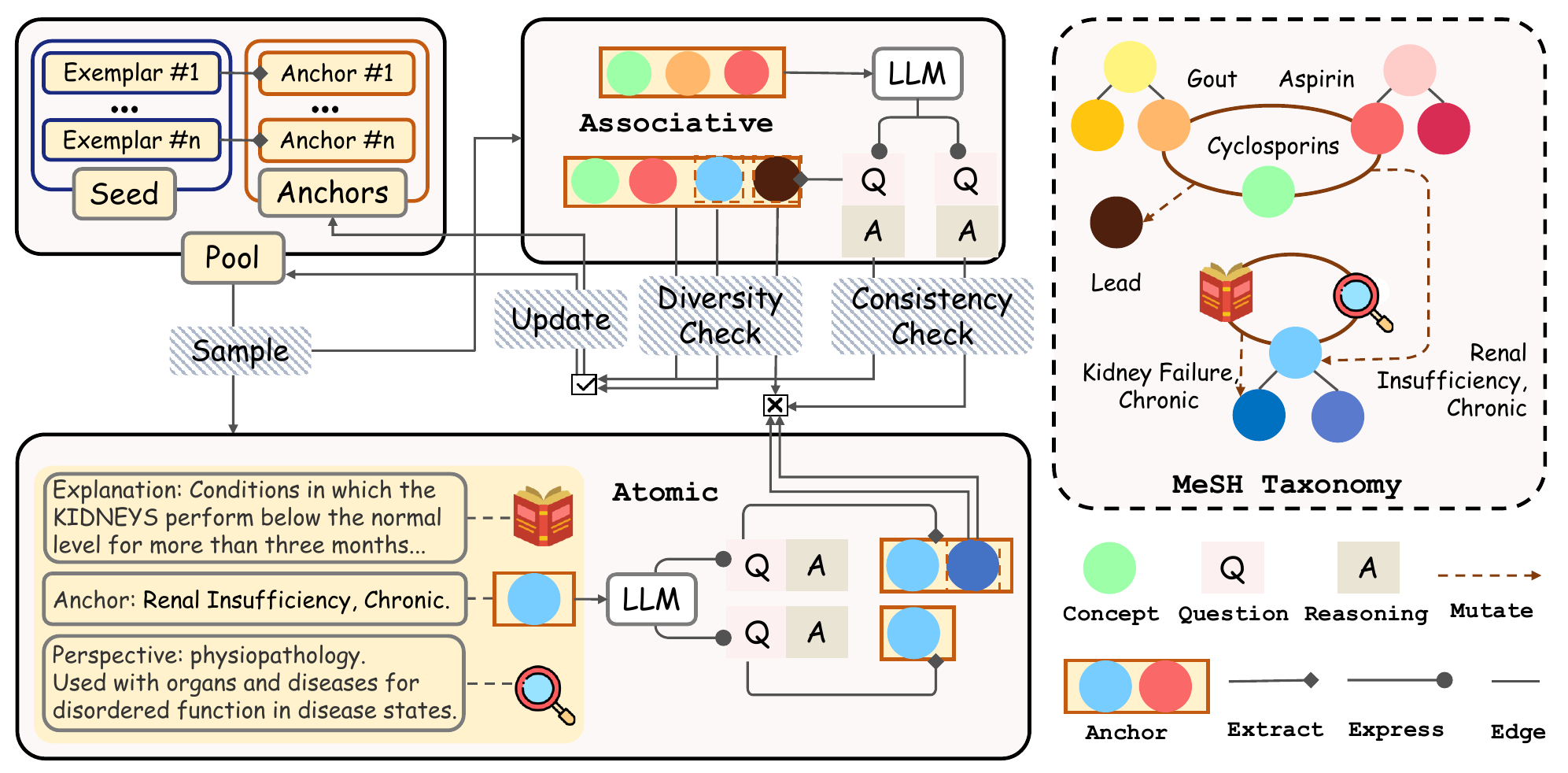}
    \vspace{-.25em}
    \caption{Overview of the SAGE framework. MeSH anchors guide dual-track generation, while consistency and diversity checks select QA instances whose projected concepts update the anchor population across epochs.}
    \vspace{-.5em}
    \label{fig:overview}
\end{figure*}

To support both concept-focused grounding and relation-oriented exploration, as illustrated in Figure~\ref{fig:overview}, we route each anchor configuration according to its cardinality. Singleton configurations are assigned to atomic generation, whereas multi-concept configurations are assigned to associative generation.

Let $\mathbf{A} \in \mathcal{S}^{(t)}$ be the target anchor configuration for the current generation step. The generation of a candidate instance $x$ is defined as follows:
\begin{equation}
    x \sim
    \begin{cases}
        P_{\text{atomic}}(x \mid \{c\}, \mathcal{K}_c)
        & \text{if } \mathbf{A}=\{c\}, \\
        P_{\text{assoc}}(x \mid \mathbf{A}, \mathcal{E}_{\text{few}}^{(t)})
        & \text{if } |\mathbf{A}|>1
    \end{cases}
\end{equation}
where $P_{\text{atomic}}$ governs single-concept synthesis and $P_{\text{assoc}}$ governs multi-concept synthesis.

\noindent\textbf{Atomic Generation.}
This track processes singleton anchor configurations ($|\mathbf{A}| = 1$), including newly surfaced concepts introduced during anchor evolution. Let $c$ denote the sole concept in $\mathbf{A}$. To provide explicit taxonomy-derived grounding without relying on few-shot demonstrations, we construct a taxonomy-derived context $\mathcal{K}_c$.
Specifically, we retrieve the authoritative definition $\delta(c)$ and the set of allowed qualifiers $\mathcal{Q}(c)$ (e.g., pathology, adverse effects) from $\mathcal{T}$. We define $\mathcal{K}_c=\{\delta(c),u_i\}$, where $u_i\sim\mathcal{U}(\mathcal{Q}(c))$ is a clinical qualifier sampled for the present generation. The distribution over candidate instances is then given by:
\begin{equation}
    P_{\text{atomic}}(x \mid \{c\},\mathcal{K}_c)
=
P_{\mathcal{M}}
(
x \mid \text{Prompt}(c,\delta(c),u_i)
)
\end{equation}
To introduce controlled variation, we generate $N$ candidate instances in parallel using different qualifiers from $\mathcal{Q}(c)$ when available. This qualifier-conditioned sampling provides controlled variation around concept $c$.

\noindent\textbf{Associative Generation.}
This track processes multi-concept anchor configurations ($|\mathbf{A}|>1$) and encourages relation-oriented generation among the anchored concepts. For each synthesis step, we construct a context $\mathcal{E}_{\text{few}}^{(t)}$ composed of $n_{\text{few}}$ examples drawn from a growing pool that contains both the initial seed instances and previously accepted samples. The few-shot pool and sampled context are defined as:
\begin{equation}
    \mathcal{F}^{(t)}
    =
    \mathcal{D}_{\mathrm{seed}}
    \cup
    \bigcup_{i=0}^{t-1}
    \mathcal{D}_{\mathrm{valid}}^{(i)},
    \qquad
    \mathcal{E}_{\mathrm{few}}^{(t)}
    \subseteq
    \mathcal{F}^{(t)}.
\end{equation}
The generative model $\mathcal{M}$ is then conditioned on $\mathbf{A}$ and $\mathcal{E}_{\text{few}}^{(t)}$ to synthesize a candidate $x$ that is encouraged to connect the concepts in the anchor configuration (e.g., generating a clinical scenario linking $\mathbf{A} = \{\text{``Aspirin''}, \text{``Chronic Pain''}\}$).

\subsection{Symbolic Selection and Anchor Evolution}
\label{sec:evolution}

Each epoch comprises answer-consistency selection, symbolic projection, and semantic mutation.

\noindent\textbf{Calibration via Consistency and Diversity Checking.}
Given the computational overhead of large-scale reward models in our resource-constrained setting, we combine answer-consistency checking with anchor-level diversity control. First, we adopt Self-Consistency (SC)~\cite{wang2023selfconsistency, yu2025cot} as the candidate-selection criterion. For each candidate instance $x=(q_x,r_x,y_x)$, we generate $K$ auxiliary answers $\{y'_j\}_{j=1}^{K}$ conditioned on $q_x$. We define the answer-consistency indicator as
\begin{equation}
    \mathbb{I}(x)
    =
    \mathbbm{1}
    \left(
        y_x =
        \operatorname{MajorVote}
        \left(
            \{y'_1,\ldots,y'_K\}
        \right)
    \right).
\end{equation}
Candidates satisfying the self-consistency criterion are incorporated into the accepted set $\mathcal{D}_{\mathrm{valid}}^{(t)}$.

Second, SAGE maintains concept- and configuration-level occurrence counts to control diversity across epochs. Anchor configurations whose expansion budgets have been exhausted are excluded from further sampling. 


\noindent\textbf{Symbolic Projection and Semantic Mutation.}
Transitioning from epoch $t$ to $t+1$, we construct the next active anchor population $\mathcal{S}^{(t+1)}$ by combining retained output configurations with singleton anchors instantiated from newly surfaced concepts.

First, we collect the projected anchor configurations of the accepted samples:
\begin{equation}
\mathcal{S}_{\text{proj}}^{(t)}
=
\left\{
\phi_{\mathcal{T}}(q_x)
\mid
x\in\mathcal{D}_{\mathrm{valid}}^{(t)},
\ \phi_{\mathcal{T}}(q_x)\neq\emptyset
\right\}.
\end{equation}

Second, during generation (e.g., discussing Aspirin for chronic pain), the model may introduce new, contextually relevant medical concepts absent from the input anchors  configuration (e.g., naturally surfacing the side effect $\text{``Gastric Ulcer''}$). Let $\mathbf{A}_{\text{in}}$ denote the input anchor configuration used to generate candidate instance $x$, and define the newly discovered concepts $\Delta(x)$ as:
\begin{equation}
    \Delta(x)
    =
    \phi_{\mathcal{T}}(q_x)
    \setminus
    \mathbf{A}_{\mathrm{in}}.
\end{equation}
These newly surfaced concepts are recognized through projection onto the MeSH concept space. Each new concept is additionally instantiated as a singleton anchor configuration for the subsequent epoch, providing an opportunity for dedicated Atomic Generation. The resulting singleton mutation configurations are collected as:
\begin{equation}
    \mathcal{S}_{\text{mut}}^{(t)} = \left\{ \{c\} \mid c \in \bigcup_{x \in \mathcal{D}_{\text{valid}}^{(t)}} \Delta(x) \right\}
\end{equation}

Finally, the active anchor population for the next epoch is obtained by combining retained projected configurations with singleton mutation configurations: $\mathcal{S}^{(t+1)} = \mathcal{S}_{\text{proj}}^{(t)} \cup \mathcal{S}_{\text{mut}}^{(t)}$.
Newly surfaced concepts are additionally instantiated as singleton anchor configurations, allowing them to receive dedicated taxonomy-derived grounding through Atomic Generation in subsequent epochs. Meanwhile, accepted output configurations are propagated to support continued associative exploration. This update expands the explored region of the MeSH concept space while keeping the generation trajectory connected to the fixed MeSH scaffold.

\section{Experiments}


\subsection{Experimental Setup}
To reflect resource-constrained on-premise settings, we use Qwen3-4B~\cite{yang2025qwen3} with thinking disabled as the common data-synthesis backbone for all methods. To assess the quality of each resulting dataset, we fine-tune a fresh Qwen3-4B-Base model using an identical supervised fine-tuning protocol. Unless otherwise specified, all trainee models are fine-tuned for 5 epochs with a learning rate of $1\times10^{-5}$ and a global batch size of 16. Further configurations are provided in Appendix~\ref{hyper}.

\subsection{Baselines and Benchmarks}
We compare SAGE against three categories of baselines:
(1) \textbf{Self-derived synthesis methods}: Self-Instruct~\cite{wang2023self}, CoT-Self-Instruct~\cite{yu2025cot}, ScaleQuest~\cite{ding2025unleashing}, KPDDS~\cite{huang2025key}, and MathScale~\cite{tang2024mathscale}. To ensure fairness, we generate 10k samples for all methods using the same seed set.
(2) \textbf{Document-based methods}: Easy-Dataset~\cite{miao2025easy}, Genie~\cite{yehudai2024genie}, Self-QA~\cite{zhang2023self}, and Wrap~\cite{maini2024rephrasing}. We use MeSH ontology scope notes as the unified reference corpus for these baselines.
(3) \textbf{High-quality open-source datasets}: m1k, m23k~\cite{huangM1UnleashPotential2025}, Huatuogpt-o1-SFT~\cite{chen2025towards}, and MedReason33k~\cite{wuMedReasonElicitingFactual2025}.

\begin{table*}[!htbp]
\centering
\resizebox{\textwidth}{!}{%
\begin{tabular}{llccccccc}
\toprule
\textbf{Model} & \textbf{Method/Dataset} & \textbf{Scale} & \textbf{AVG} & \textbf{MedMCQA} & \textbf{MedQA} & \textbf{PubMedQA} & \textbf{MMLU-Pro} & \textbf{GPQA} \\
 & & & & & & & \textbf{(med)} & \textbf{(med)} \\
\midrule
Qwen3-4B-Base & Few-shot Prompting & NA & 26.15 & 29.14 & 29.69 & 35.20 & 14.66 & 22.05 \\
Qwen3-4B NoThinking & NA & NA & 62.54 & 56.78 & 63.63 & 71.00 & 64.63 & 56.67 \\
Qwen3-4B Thinking & NA & NA & \underline{66.18} & 59.81 & 72.03 & 72.60 & 71.60 & 54.87 \\
\midrule
\multicolumn{9}{l}{\textit{\textbf{Open Source Datasets}}} \\
Qwen3-4B-Base & m1k~\citeyearpar{huangM1UnleashPotential2025} & 1k & 63.40 & 59.26 & 72.19 & 74.20 & 67.49 & 43.85 \\
Qwen3-4B-Base & m23k~\citeyearpar{huangM1UnleashPotential2025} & 23k & \underline{67.39} & 64.19 & 76.20 & 76.00 & 68.79 & 51.79 \\
Qwen3-4B-Base & Huatuo-SFT~\citeyearpar{chen2025towards} & 20k & 52.80 & 54.08 & 60.80 & 56.70 & 53.68 & 38.72 \\
Qwen3-4B-Base & MedReason~\citeyearpar{wuMedReasonElicitingFactual2025} & 33k & 58.60 & 53.60 & 60.41 & 76.90 & 63.13 & 38.97 \\
\midrule
\multicolumn{9}{l}{\textit{\textbf{Self-derived Method}}} \\
Qwen3-4B-Base & Self-Instruct~\citeyearpar{wang2023self} & 10k & 53.66 & 52.35 & 50.12 & 69.70 & 53.29 & 42.82 \\
Qwen3-4B-Base & ScaleQuest~\citeyearpar{ding2025unleashing} & 10k & 62.91 & 57.42 & 67.16 & 76.80 & 64.95 & 48.21 \\
Qwen3-4B-Base & KPDDS~\citeyearpar{huang2025key} & 10k & 58.99 & 52.52 & 61.74 & 71.60 & 62.15 & 46.92 \\
Qwen3-4B-Base & MathScale~\citeyearpar{tang2024mathscale} & 10k & 57.85 & 55.01 & 58.92 & 73.70 & 60.85 & 40.77 \\
Qwen3-4B-Base & CoT-Self-Instruct~\citeyearpar{yu2025cot} & 10k & \underline{63.71} & 62.08 & 67.40 & 70.00 & 61.89 & 57.18 \\
\midrule
\multicolumn{9}{l}{\textit{\textbf{Document-based Method}}} \\
Qwen3-4B-Base & Easy-Dataset~\citeyearpar{miao2025easy} & 28k & 49.11 & 46.90 & 44.07 & 70.30 & 48.40 & 35.90 \\
Qwen3-4B-Base & Genie~\citeyearpar{yehudai2024genie} & 30k & 49.98 & 49.32 & 51.77 & 63.90 & 48.99 & 35.90 \\
Qwen3-4B-Base & Self-QA~\citeyearpar{zhang2023self} & 230k & 46.34 & 48.98 & 51.06 & 48.90 & 47.10 & 35.64 \\
Qwen3-4B-Base & Wrap~\citeyearpar{maini2024rephrasing} & 98K & \underline{50.20} & 47.53 & 47.13 & 69.50 & 47.10 & 39.74 \\
\midrule
\multicolumn{9}{l}{\textit{\textbf{Ours}}} \\
Qwen3-4B-Base & Ours, w/o Iteration & 10k & 66.45 & 64.24 & 68.26 & 80.20 & 60.33 & 59.23 \\
Qwen3-4B-Base & Ours, w/o Parallel & 10k & 66.81 & 64.04 & 67.95 & 82.80 & 60.78 & 58.46 \\
Qwen3-4B-Base & Ours, w/o Atomic gen. & 10k & 66.21 & 65.29 & 66.69 & 77.20 & 61.89 & 60.00 \\
Qwen3-4B-Base & Ours, w/o Associative gen. & 10k & 66.13 & 65.24 & 69.84 & 77.90 & 62.02 & 55.64 \\
Qwen3-4B-Base & \textbf{Ours (Full)} & 10k & \textbf{68.12} & 64.98 & 70.07 & 80.80 & 63.19 & 61.54 \\
\bottomrule
\end{tabular}%
}
\caption{Performance of models trained on datasets generated or collected by various paradigms, evaluated on mainstream medical QA benchmarks. The highest average score within each paradigm is \underline{underlined}. The performance of our method is highlighted in \textbf{bold}. Units are in percentage (\%). }
\vspace{-.5em}
\label{tab:main_results}
\end{table*}

For evaluation, we utilize a suite of ten medical question-answering benchmarks that probe different aspects of medical reasoning and generalization. Following~\citet{chen2025towards}, our suite includes MedMCQA~\cite{pal2022medmcqa}, MedQA~\cite{jin2021disease}, and PubMedQA~\cite{jin2019pubmedqa}, which focus on domains well-aligned with our seed data. To measure cross-domain generalization, we employ the medical subset of GPQA~\cite{rein2024gpqa} and MMLU-Pro~\cite{wang2024mmlu}. For robustness and advanced reasoning, we use high-difficulty benchmarks: Lancet\footnote{\url{https://www.thelancet.com/}}, NEJM\footnote{\url{https://www.nejm.org/}}, MedBullets~\cite{chen2025benchmarking}, and MedXpertQA~\cite{zuo2025medxpertqa}. Accuracy is adopted as the evaluation metric for all tasks following previous works~\cite{chen2025towards,huangM1UnleashPotential2025,yu2025finemedlmo,wang2025baichuan}.


\subsection{Main Results}
\label{sec:main_results}

As shown in Table~\ref{tab:main_results}, SAGE consistently outperforms baseline methods. Compared to self-derived approaches, SAGE achieves substantial gains by anchoring generation in factual knowledge. It also surpasses document-based methods, suggesting that our approach effectively integrates atomic facts into coherent scenarios rather than merely extracting them; we provide concrete case studies illustrating this distinction in Appendix~\ref{ap:baseline_case}. Notably, SAGE matches or exceeds the performance of large-scale datasets and models fine-tuned on massive CoT data. This efficiency highlights the value of our evolutionary strategy, which maximizes information density by balancing concept mastery with scenario-based reasoning.

\subsection{Ablation Study}

\noindent\textbf{Contribution of Evolutionary Components.}

\begin{figure}[ht]
    \centering
    \includegraphics[width=.85\linewidth]{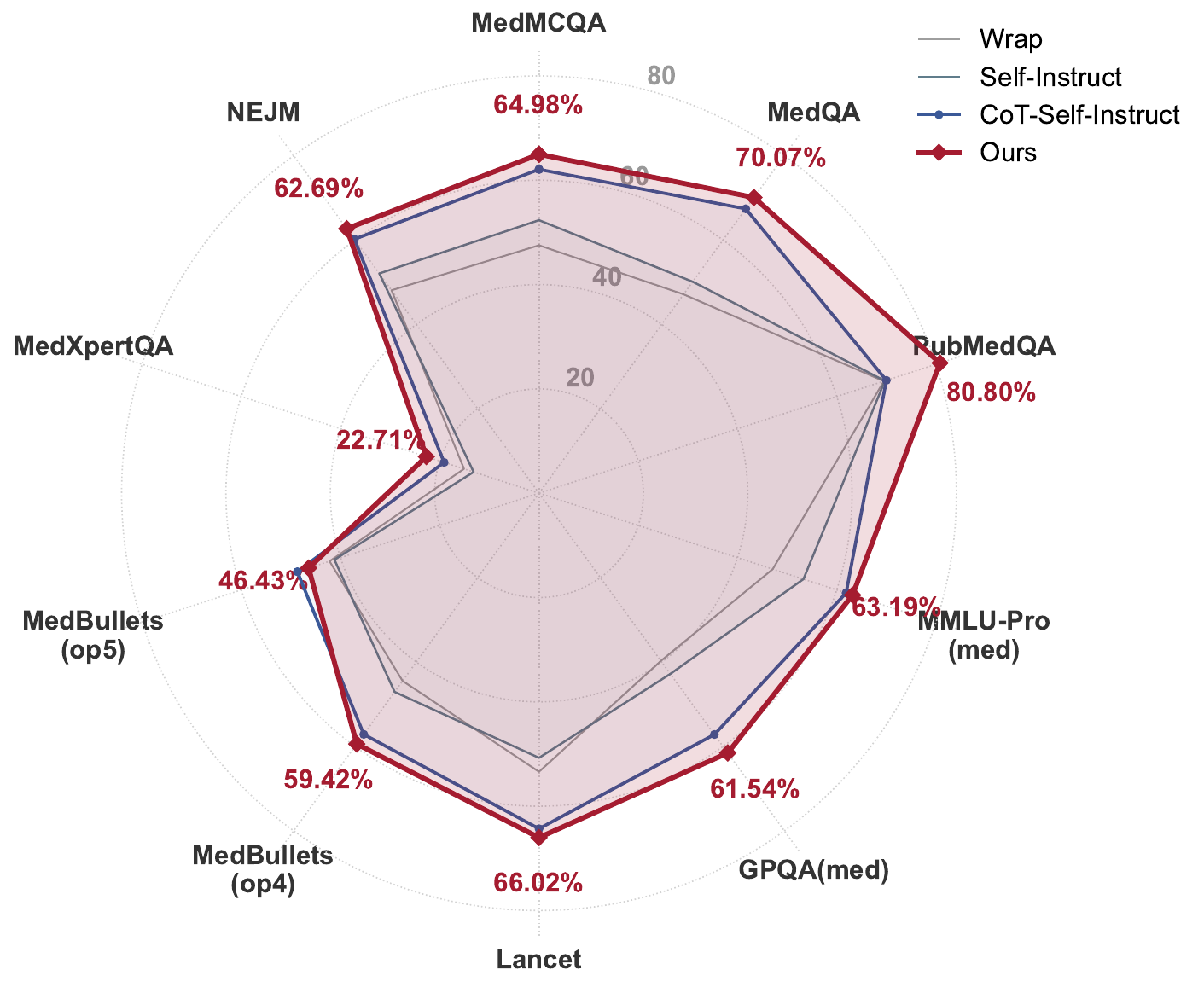}
    \caption{Comparison between our method and baselines across diverse medical QA benchmarks, including out-of-distribution tasks.}
    \vspace{-.5em}
    \label{fig:radar}
\end{figure}

We conducted comprehensive ablation studies on all four components of our proposed method; the results are summarized in Table~\ref{tab:main_results}, alongside extended validations of our mapping function, corpus selection, and factual filtering strategies (Appendices~\ref{ap:mapping_evaluation}, \ref{ap:corpus_ablation}, \ref{ap:filtering}).

Specifically, for the \textit{w/o Iteration} variant, the target dataset size was achieved by increasing multi-sample generation within a single iteration, without leveraging the parametric knowledge of the LLM to evolve the concept graph across successive iterations.

\begin{figure}[ht]
    \centering
    \includegraphics[width=0.85\linewidth]{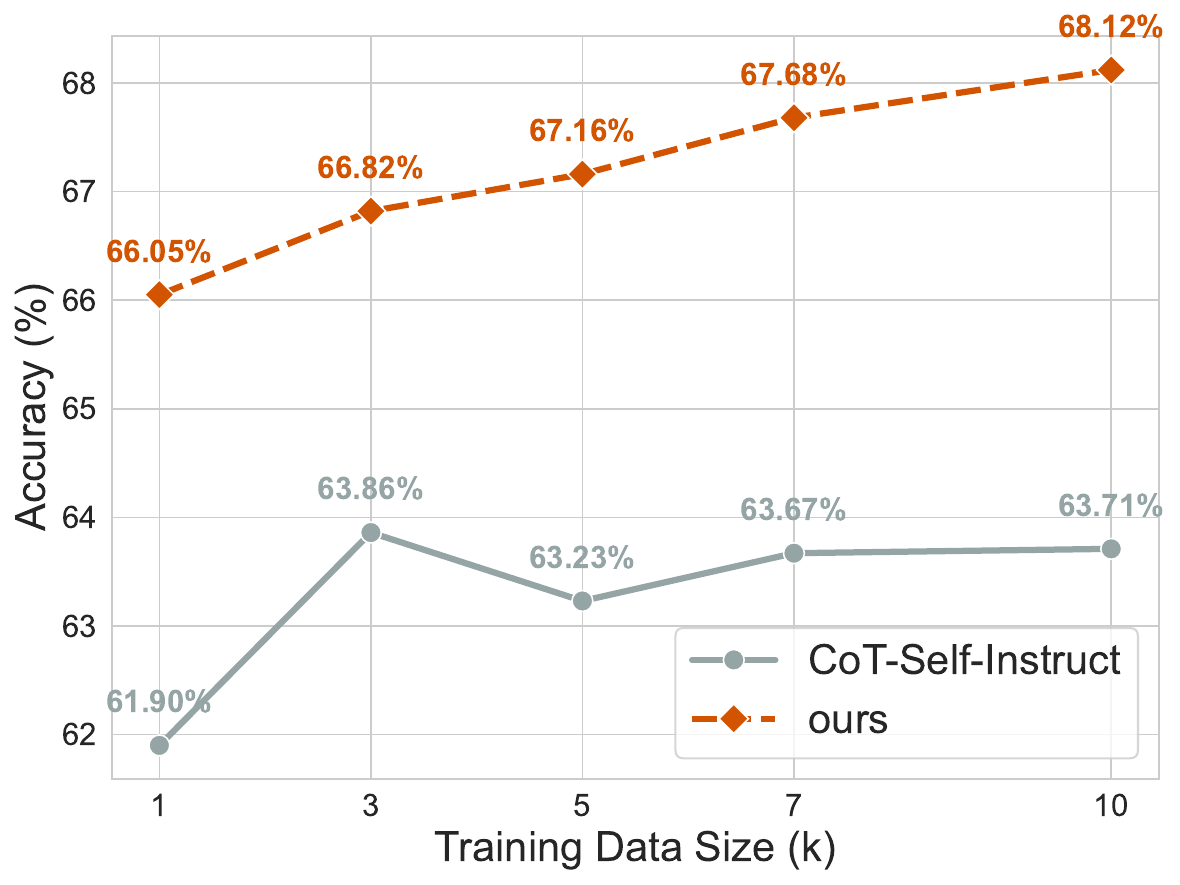}
    \caption{Performance scaling analysis comparing our method with CoT-Self-Instruct across different training data sizes from 1k to 10k samples.}
    \label{fig:scaling}
    \vspace{-.5em}
\end{figure}

\begin{figure*}[]
    \centering
    \includegraphics[width=0.95\linewidth]{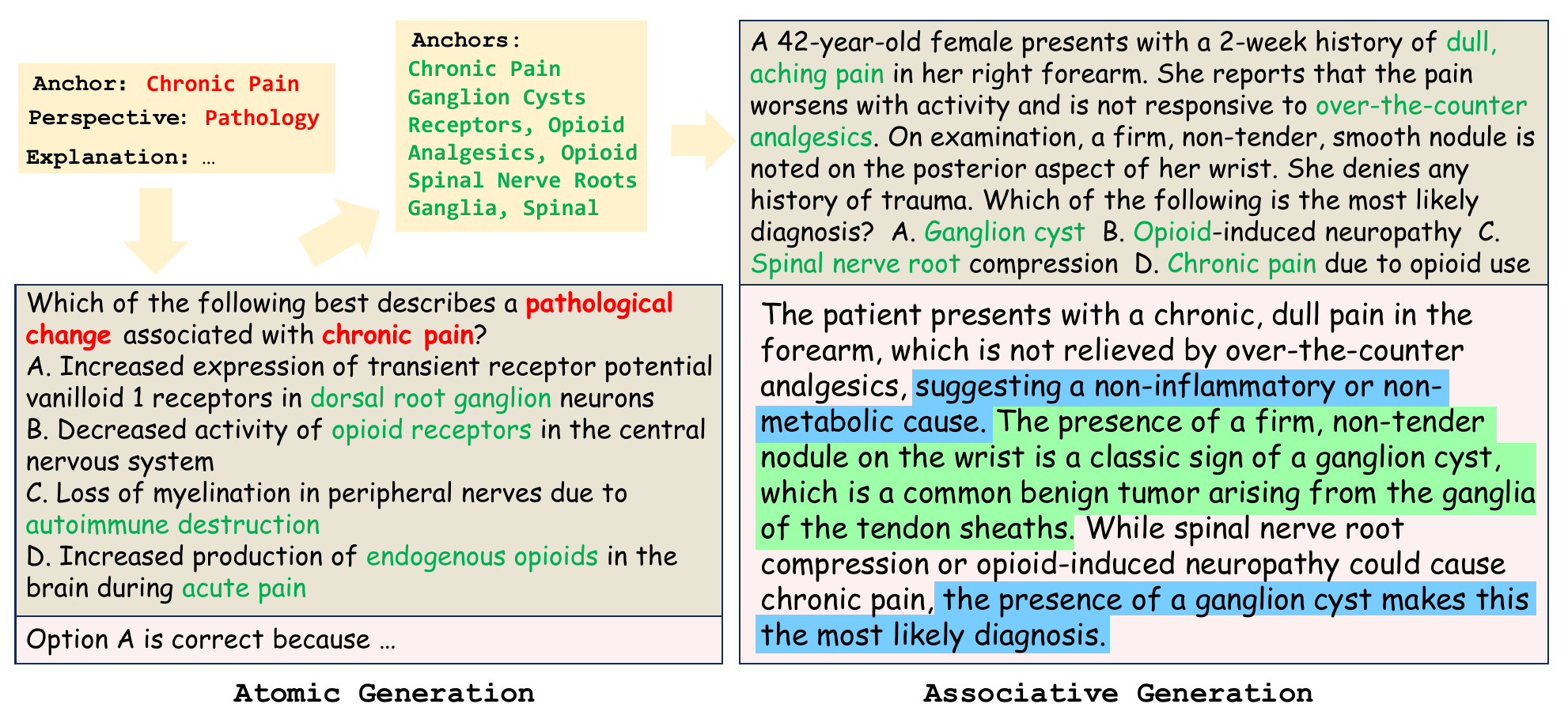}
    \caption{A detailed case on dual-track evolutionary synthesis.}
    \label{fig:at_and_as_case}
    \vspace{-.75em}
\end{figure*}

For the \textit{w/o Parallel} variant, the target sample size was obtained by relaxing the criteria for semantic anchor deduplication in subsequent iterations.
Furthermore, versions lacking either of the two question types consistently exhibited marked performance degradation, underscoring the significance of maintaining diversity in question formats.

\noindent\textbf{Scalability and Efficiency.}

Figure~\ref{fig:scaling} illustrates our method's high data efficiency, maintaining strong performance across varying scales and rivaling much larger parameter models (Appendix~\ref{ap:ceiling}). Token usage analysis (Appendix~\ref{ap:efficiency}) reveals it consumes significantly fewer tokens than baselines, confirming SAGE's superior information density. Additionally, our offline semantic mapping ensures extreme computational efficiency compared to standard online API tools (Appendix~\ref{ap:mapping_evaluation}).

\noindent\textbf{Generalization and Robustness.}
As shown in Figure~\ref{fig:radar}, our method consistently outperforms baselines on out-of-distribution (OOD) QA benchmarks (Appendix~\ref{ap:extended_benchmark}). Crucially, this strong generalization extends beyond QA to other clinical tasks like MedNLI and MTS-Dialog (Appendix~\ref{ap:nonqa}). Furthermore, ablation studies across various generator models and architectures (Appendix~\ref{ap:model_ablation}) confirm the robustness of our approach irrespective of the underlying model.

\section{Discussion}

\subsection{Qualitative Case: Mitigating Hallucination and Enhancing Reasoning.}
\label{sec:qual_case}

While document-based data synthesis methods can generate factually correct content, they often produce questions that assess only simple concept recall, rather than complex reasoning or the application of knowledge to specific clinical scenarios. This limitation significantly diminishes the learning value of the resulting training corpus. In contrast, self-derived synthesis methods can create corpora that emphasize reasoning, but typically lack mechanisms for factual verification or refinement. This often results in severe hallucinations, which are unacceptable in high-risk fields such as medicine. A detailed qualitative comparison of these limitations is provided in Appendix~\ref{ap:baseline_case}.

We begin with a single medical concept serving as a semantic anchor, accompanied by explanatory text for knowledge grounding. Using a specific examination perspective, we perform atomic generation to produce question-answer pairs that consolidate the model's factual understanding of the given concept. Subsequently, leveraging the LLM's capacity for self-association, we extract multiple concepts involved in the generated questions. These extracted concepts then serve as semantic anchors for the next phase of associative generation, enabling the creation of training materials that strengthen the model's ability to apply concepts in complex medical scenarios. A concrete example is illustrated in Figure~\ref{fig:at_and_as_case}.

As shown in the figure, starting from the core concept ``Chronic Pain'' with ``pathology'' as the examination perspective, the LLM generates questions that naturally introduce related concepts such as ``Spinal Nerve Roots'' and ``Ganglia, Spinal''. In the subsequent associative generation phase, these concepts are used as scaffolds to construct appropriate clinical scenarios, design educationally valuable questions, and provide responses that integrate accurate knowledge with advanced reasoning. Unlike the baselines, this two-stage process effectively mitigates the hallucination risks of self-derived methods while overcoming the shallow reasoning limitations inherent in document-based approaches.




\begin{figure}[h]
    \centering
    \includegraphics[width=0.95\linewidth]{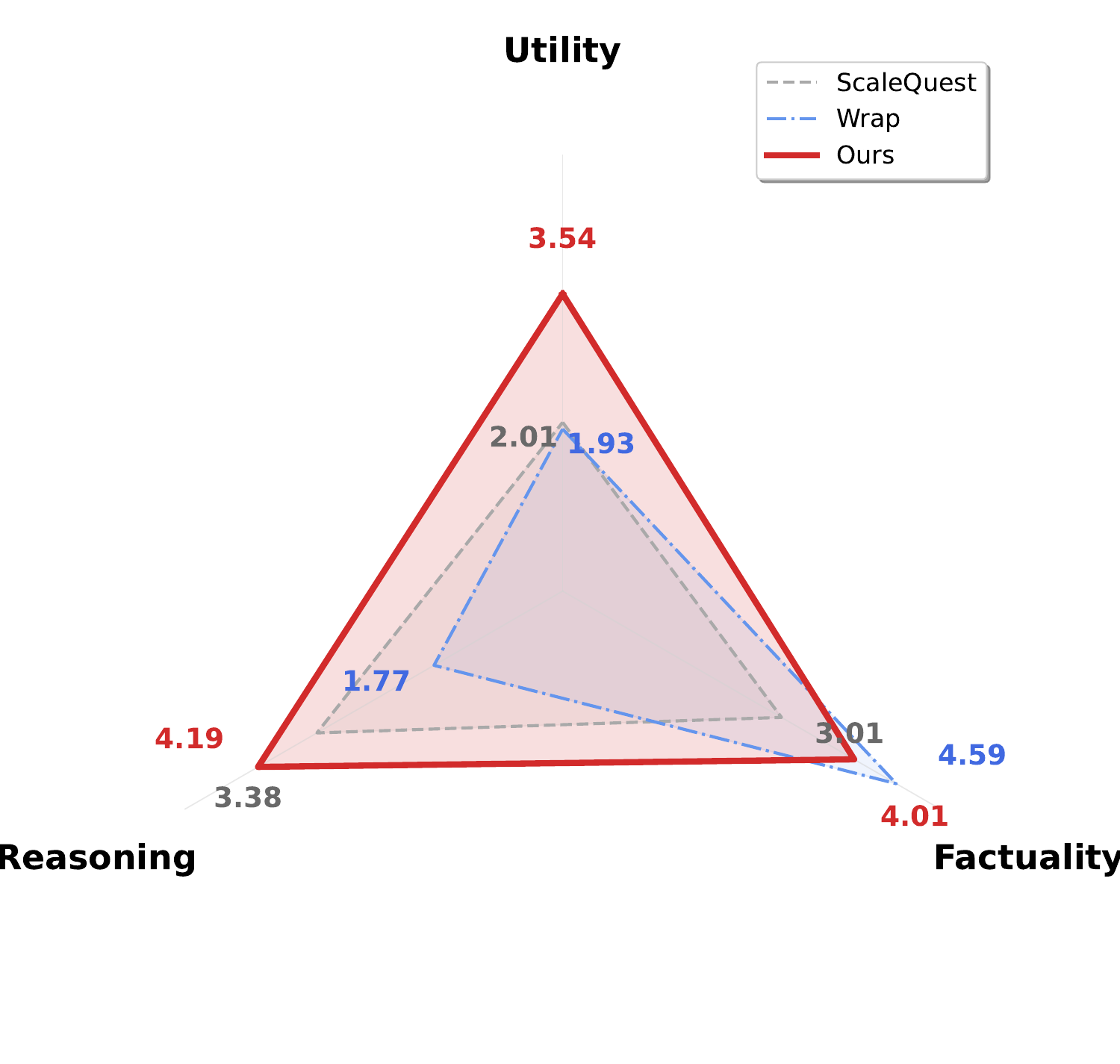}
    \vspace{-1em}
    \caption{Comparison of different methods across factuality, reasoning, and utility.}
    \vspace{-.5em}
    \label{fig:judge_radar}
\end{figure}

\subsection{Quantitative Assessment: Factuality, Reasoning, and Utility.}
We evaluated 1{,}000 generated samples using DeepSeek-V3.2~\cite{guo2025deepseek} across medical factuality, reasoning, and utility (Figure~\ref{fig:judge_radar}). The document-based Wrap~\cite{maini2024rephrasing} achieves high factuality but lacks reasoning depth (consistent with Figure~\ref{fig:two_baselines_case}), limiting its utility. Conversely, the self-derived ScaleQuest~\cite{ding2025unleashing} shows moderate reasoning but suffers from factual hallucinations. SAGE achieves a superior balance of both. To ensure our factuality claims extend beyond LLM-as-a-judge, we further validated SAGE on standard hallucination benchmarks (TruthfulQA~\cite{lin2022truthfulqa} and MedHALT~\cite{pal2023med}), where it consistently minimized confabulations compared to all baselines (Appendix~\ref{ap:hallucination}).

\begin{figure}[h]
    \centering
    \includegraphics[width=0.97\linewidth]{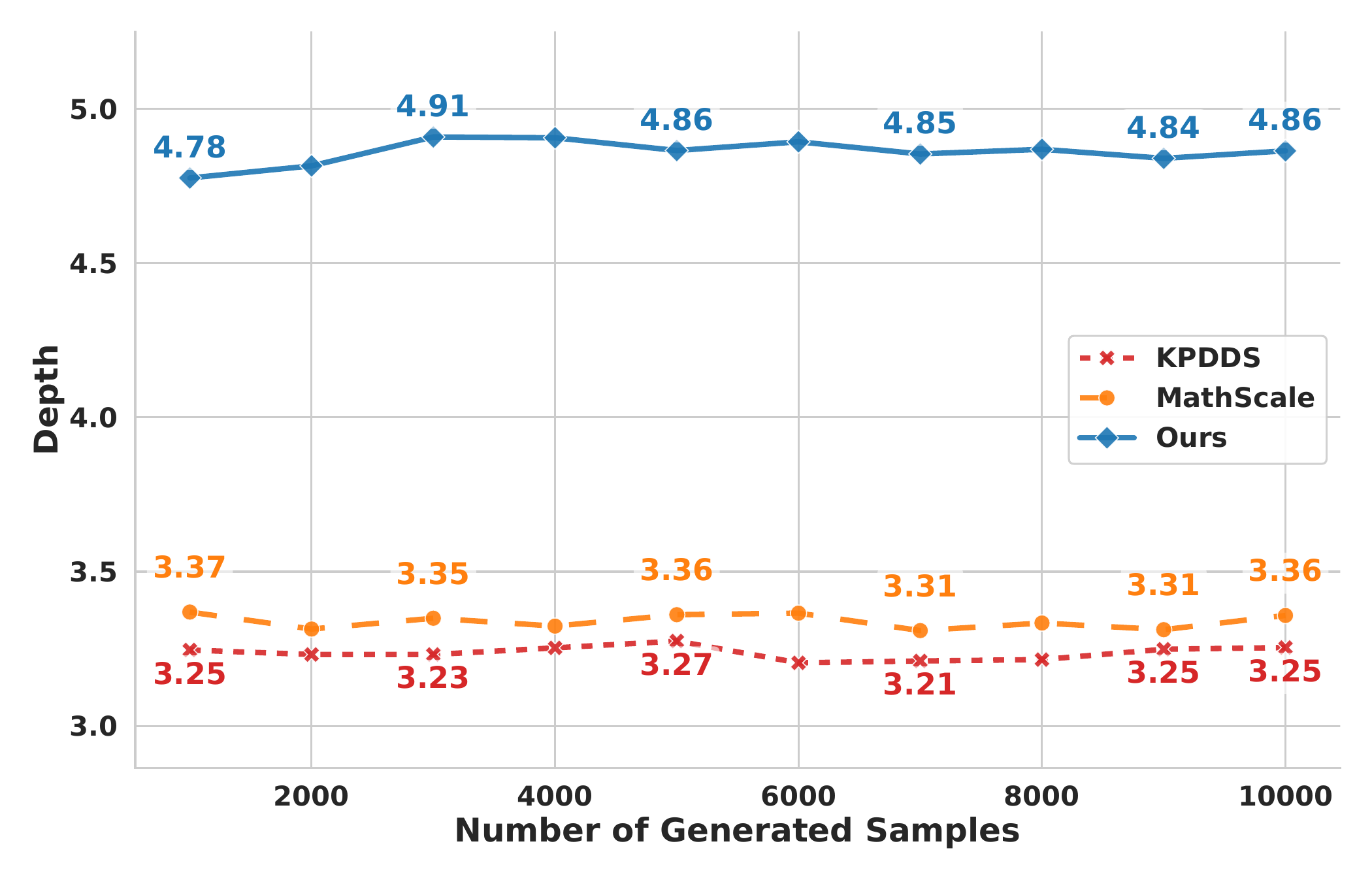}
    \caption{Comparison of semantic depth (MeSH Tree) of generated concepts during the generation process. }
    \label{fig:depth}
    \vspace{-.5em}
\end{figure}

\subsection{Combinatorial Validity and Diversity.}

\begin{figure}[h]
    \centering
    \includegraphics[width=0.97\linewidth]{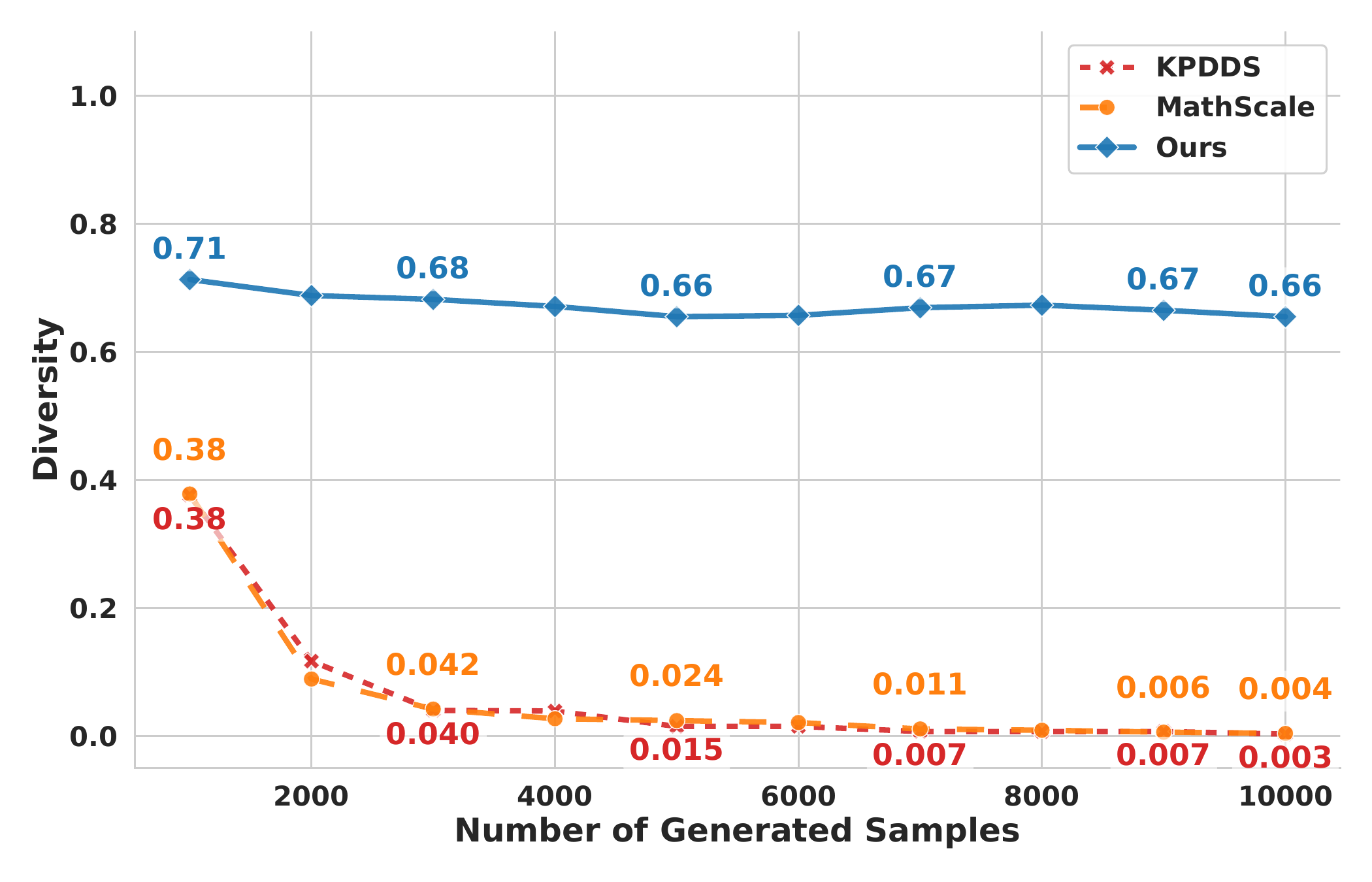}
    \caption{Comparison of dynamic diversity with increasing generated samples. Diversity is measured by the discovery rate $R = \frac{N_{new}}{B}$, where $N_{new}$ denotes the number of unique unseen concept combinations, and $B$ is the batch size.}
    \vspace{-.5em}
    \label{fig:diversity}
\end{figure}

We conducted a detailed investigation of two self-derived methods that also generate data based on concepts. Specifically, the concepts autonomously generated by the LLM in KPDDS and MathScale were mapped to specific concepts within the MeSH ontology using a medical embedding model~\cite{deka2022improved}. For each method, we generated a total of 10{,}000 data points and calculated two metrics for every batch of 1{,}000 samples: the average semantic depth of the concepts used and the diversity of concept combinations, as shown in Figures~\ref{fig:depth} and~\ref{fig:diversity}.

As illustrated, with increasing data scale, our method consistently maintains high levels of both semantic depth and combinatorial diversity. In contrast, the baseline methods, which rely on the LLM's self-directed concept generation, tend to produce broad, shallow, and frequently recurring topical concepts. Subsequent data regeneration based on statistical co-occurrence patterns within these limited initial batches fails to overcome common concept pairings, resulting in a rapid collapse in diversity.

\section{Conclusion}

We introduced SAGE, a Semantic Anchor-Guided Evolution framework that mitigates the data scarcity bottleneck for medical LLMs under strict privacy constraints. By grounding atomic and associative generation in public taxonomies, SAGE enables small local models to autonomously produce pedagogically structured data. Results show it outperforms prevalent paradigms without relying on massive corpora, establishing a practical pathway for developing reliable medical AI assistants directly from a compact knowledge base.



\section*{Limitations}

While SAGE provides a resource-efficient foundation for grounded QA data synthesis, its current design still leaves several limitations for future work. The effectiveness of the framework is partly tied to the scope and organization of external taxonomies such as MeSH, and applying SAGE to domains without mature semantic resources may require constructing task-specific knowledge schemas. Although structured anchoring helps reduce hallucination, the generated scenarios are still shaped by the combinatorial coverage of anchor concepts and may not fully reflect the nuanced, open-ended reasoning patterns present in authentic clinical narratives. In addition, due to the scale of the synthesized data, we did not conduct comprehensive clinician double-checking for all generated samples, which may leave some subtle clinical inaccuracies unexamined. Our evaluation also primarily focuses on question-answering benchmarks, while broader validation on interactive clinical tasks, such as patient dialogue simulation or diagnostic decision support, would further clarify the practical utility of the framework. Future directions include dynamically expanding semantic anchors, integrating multimodal medical knowledge such as imaging data, and adapting SAGE to other high-stakes, knowledge-intensive domains beyond medicine.

\section*{Ethics Statement}
This research utilizes only publicly available, de-identified resources: the structured MeSH taxonomy and open-source medical QA datasets are used exclusively for research purposes. All training data are synthetically generated, circumventing privacy and confidentiality concerns associated with real patient information. No human subjects, personal health data, or protected clinical records were involved in this study. Nonetheless, we acknowledge that models trained on synthetic data may reproduce or amplify biases present in the underlying taxonomies or generation processes, and their outputs should not be interpreted as clinical advice. Any real-world deployment of this methodology or its derivatives must undergo rigorous clinical validation, adhere to all relevant regulatory standards, and be implemented in human-supervised healthcare workflows.

\section*{Acknowledgements}
This work was supported by the Guizhou Provincial Program on Commercialization of Scientific and Technological Achievements (Qiankehezhongyindi [2025] No. 006).

\bibliography{custom}

\newpage
\appendix

\section{Data and Knowledge Base}

\subsection{Knowledge Base Description}
\label{ap:knowledge_base}
Medical Subject Headings (MeSH) is a comprehensive controlled vocabulary developed by the National Library of Medicine (NLM)\footnote{\url{https://www.nlm.nih.gov/}}, primarily used for indexing, cataloging, and searching biomedical and health-related information. As the authoritative thesaurus for the MEDLINE\footnote{\url{https://www.nlm.nih.gov/medline/medline_home.html}} and PubMed\footnote{\url{https://pubmed.ncbi.nlm.nih.gov/}} databases, MeSH enforces uniform terminology and consistency in the retrieval of biomedical literature. Its hierarchical tree structure organizes descriptors from broad, general categories down to specific concepts; each descriptor is assigned one or more ``Tree Numbers'' that indicate its precise location within this hierarchy. This organization robustly encodes semantic relationships between medical concepts, making MeSH an essential foundation for constructing domain-specific knowledge graphs and ontologies.
We leverage the built-in synonym information within the MeSH ontology to implement the mapping function $\phi$, as described in Section~\ref{sec:method}, using a lightweight and efficient string matching approach. To focus on medically relevant content, we retain only mid-to-deep level terms from core medical categories as candidate semantic anchors(including Anatomy; Diseases; Chemicals and Drugs; Analytical, Diagnostic and Therapeutic Techniques, and Equipment; Psychiatry and Psychology; and Phenomena and Processes).  Furthermore, when selecting a single anchor and its associated inquiry perspective, we consider only the core medical perspectives pertinent to that concept, such as clinical diagnosis and treatment, pharmacology, pathophysiology, and mechanism-based inquiry.

\subsection{Seed Dataset Overview}
We utilize the m1k~\cite{huangM1UnleashPotential2025} dataset as the seed for our entire data synthesis pipeline. It provides concepts distributions and few-shot examples that serve as initial scaffolds for guiding small-scale models in complex data generation tasks.

The m1k dataset represents a curated medical QA benchmark, distilled from an initial pool of 196K public samples across MedMCQA, MedQA-USMLE, HeadQA, and PubMedQA. The construction process involved strict difficulty filtering, isolating 37K complex questions that Qwen2.5-7B and Qwen2.5-32B failed to solve. These were subsequently augmented with verified Chain-of-Thought (CoT) solutions from DeepSeek-R1. Ultimately, through diversity sampling stratified by MeSH categories and balanced by source, a core training set of 1,000 instances was established.

\subsection{Benchmark Datasets}
To comprehensively assess the efficacy of our proposed method, we utilize a suite of established medical QA benchmarks:

\begin{itemize}
    \item \textbf{MedMCQA}~\cite{pal2022medmcqa}: A massive dataset of over 194,000 multiple-choice questions covering 21 medical disciplines and 2,400 topics. Originated from real-world Indian entrance exams (AIIMS and NEET-PG), we utilize the specific 4,183-question test set adopted by~\cite{chen2025towards} (Apache-2.0).
    \item \textbf{MedQA}~\cite{jin2021disease}: Derived from USMLE practice materials, this corpus contains 12,723 questions compiled from 18 clinical textbooks. Solving these requires expert-level reasoning and the ability to synthesize evidence from multiple documents. We employ the 1,273-sample test split following~\cite{chen2025towards} (CC-BY-4.0).
    \item \textbf{PubMedQA}~\cite{jin2019pubmedqa}: A biomedical QA task grounded in PubMed abstracts. It features 1,000 expert-labeled multiple-choice questions drawn from a broader collection of 211k articles (MIT).
    \item \textbf{MMLU-Pro (Medical)}~\cite{wang2024mmlu}: The Professional Medicine segment of the Massive Multitask Language Understanding benchmark, targeting advanced clinical knowledge. We adopt the test splits used in~\cite{chen2025towards} (MIT).
    \item \textbf{GPQA (Medical)}~\cite{rein2024gpqa}: Representing the biomedical portion of the Graduate-Level QA dataset, these questions are expert-written and designed to be Google-proof, resisting simple retrieval strategies. We strictly use the biology/medical subset and splits from~\cite{chen2025towards} (CC-BY-4.0).
    \item \textbf{Lancet \& NEJM}: Two compact datasets constructed from clinical case reports published in \textit{The Lancet} and the \textit{New England Journal of Medicine} (NEJM), curated according to~\cite{huangM1UnleashPotential2025}.
    \item \textbf{MedBullets}~\cite{chen2025benchmarking}: A collection of practice items from the MedBullets platform. Following~\cite{huangM1UnleashPotential2025}, we focus on the high-difficulty subsets (Levels 4 and 5), labeled as MedBullets\_Op4 and MedBullets\_Op5, each containing approximately 100 rigorous questions.
    \item \textbf{MedXpertQA}~\cite{zuo2025medxpertqa}: A set of 50 expert-crafted questions requiring multi-step reasoning, employed here for qualitative analysis (MIT).
\end{itemize}

\section{Implementation Details and Hyperparameters}
\label{hyper}
Candidate instances are sampled with a temperature of $0.7$, top-$p$ of $0.95$, and a maximum generation length of 4,096 tokens. For each eligible anchor configuration, we generate $N=2$ candidates. Associative generation uses $n_{\mathrm{few}}=2$ examples sampled from the growing few-shot pool, whereas self-consistency checking uses $K=3$ auxiliary answers. 

\section{Training on Weaker Models}
\label{ap:model_ablation}
\begin{figure*}[!h]
    \centering
    \includegraphics[width=0.98\linewidth]{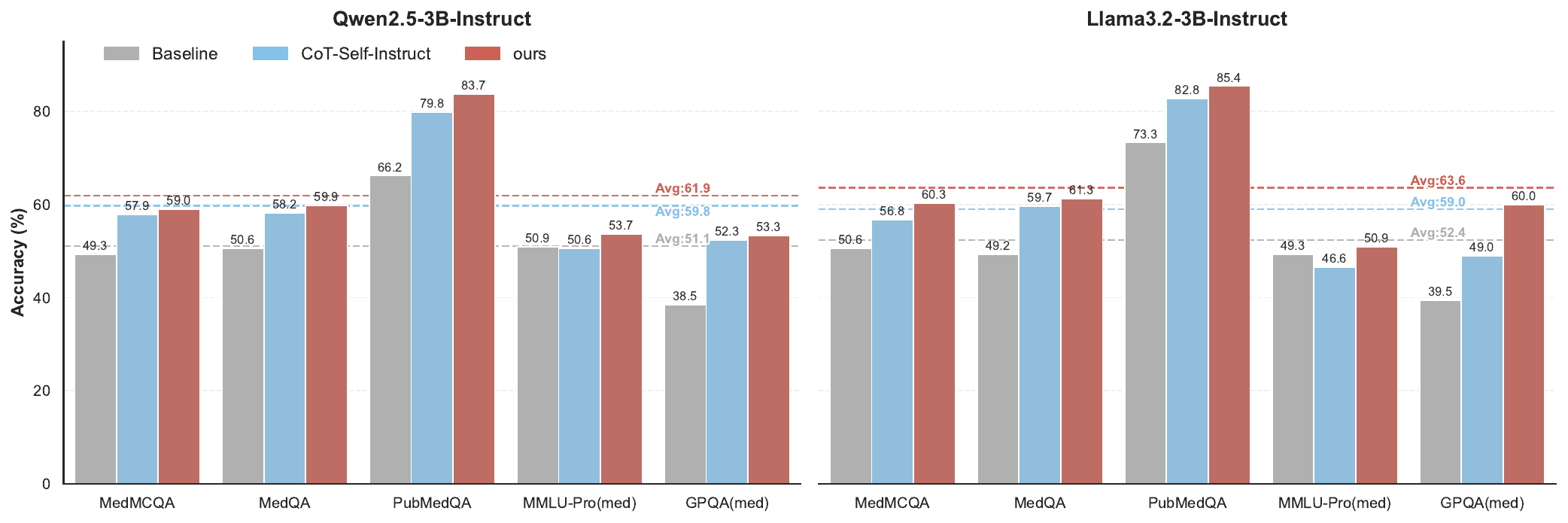}
    \caption{Performance on the benchmark after training Qwen2.5-3B-Instruct and Llama3.2-3B-Instruct.}
    \label{fig:slm}
\end{figure*}

To demonstrate the robustness of our method in real-world scenarios, we evaluated data generated from Qwen3-4B on weaker models. As shown in Figure~\ref{fig:slm}, our method achieved significant advantages across different model architectures, highlighting its strong adaptability.

\section{Generalization on non-QA Benchmark}
\label{ap:nonqa}
To demonstrate SAGE's generalization on non-QA benchmark, we further evaluated it on MedNLI \cite{shivade2017mednli} (Natural Language Inference) and MTS-Dialog \cite{abacha2023empirical} (Medical Summarization). As shown in Table \ref{tab:nonqa}, SAGE consistently outperforms synthesis baselines on both tasks. SAGE provides the strongest gains among all synthesis paradigms.

\begin{table}[]
\centering
\begin{tabular}{lcc}
\toprule
\textbf{Strategy}   & \textbf{MedNLI} & \textbf{MTS} \\
\midrule
Wrap                & 72.08           & 83.62        \\
CoT-Self-Instruct   & 84.60           & 83.16        \\
\textbf{SAGE}       & \textbf{87.27}  & \textbf{85.02} \\
\bottomrule
\end{tabular}
\caption{Generalization performance of SAGE compared to baselines. All trainee models are based on Qwen3-4B-Base. We report Accuracy for MedNLI and BERTScore for MTS-Dialog. All results are scaled to percentages (\%).}
\label{tab:nonqa}
\end{table}

\section{Performance Ceiling and Cross-Scale Comparison}
\label{ap:ceiling}

To provide context on the difficulty of the clinical benchmarks evaluated in this study and offer insight for non-specialist readers, we present a performance ceiling and a cross-scale comparison. Table \ref{tab:performance_ceiling} compares our model against recent state-of-the-art (SOTA) open-source medical LLMs, heavy-weight general models, and flagship reasoning models.

As the table illustrates, these clinical benchmarks are challenging. Even parameter-heavy models like UltraMedical-8B struggle to surpass a 63\% average. By providing this ceiling, it becomes evident that pushing a 4B model to a 68.12\% average, matching the 8B SOTA and approaching the performance of a 72B general model, is a highly non-trivial achievement. This underscores the efficiency and high quality of the SAGE synthetic data paradigm.

\begin{table*}[!h]
\centering
\resizebox{\textwidth}{!}{
\begin{tabular}{lcccccc}
\toprule
\textbf{Model} & \textbf{MedMCQA} & \textbf{MedQA} & \textbf{PubMedQA} & \textbf{MMLU-Pro} & \textbf{GPQA} & \textbf{AVG} \\
& & & & (med) & (med) & \\
\midrule
SAGE-4B                        & 64.98 & 70.07 & 80.80 & 63.19 & 61.54 & 68.12 \\
\midrule
\multicolumn{7}{c}{\textit{$\sim$8B Scale}} \\
\midrule
UltraMedical-8B \cite{zhang2024ultramedical} & 59.17 & 71.64 & 70.30 & 61.43 & 50.26 & 62.56 \\
m1-7B \cite{huangM1UnleashPotential2025}                     & 61.49 & 71.96 & 74.00 & 63.65 & 46.67 & 63.55 \\
MedReason-8B \cite{wuMedReasonElicitingFactual2025}       & 60.17 & 70.78 & 78.50 & 65.08 & 52.56 & 65.42 \\
HuatuoGPT-o1-8B \cite{chen2025towards}       & 63.78 & 75.02 & 80.50 & 64.23 & 57.18 & 68.14 \\
\midrule
\multicolumn{7}{c}{\textit{Above 70B Scale}} \\
\midrule
Qwen2.5-72B-Instruct                    & 66.60 & 74.55 & 70.89 & 66.06 & 62.05 & 68.03 \\
UltraMedical-70B \cite{zhang2024ultramedical}& 72.94 & 83.90 & 80.00 & 73.94 & 58.72 & 73.90 \\
\midrule
\multicolumn{7}{c}{\textit{Ceiling Reference}} \\
\midrule
Deepseek-V3-0324                        & 75.30 & 88.66 & 69.33 & 84.67 & 66.00 & 76.79 \\
Deepseek-R1-0528                        & 80.20 & 93.07 & 68.32 & 90.10 & 73.27 & 80.99 \\
\bottomrule
\end{tabular}
}
\caption{Cross-scale comparison and performance ceiling on clinical benchmarks. Results are reported in percentages (\%). SAGE-4B matches the performance of SOTA 8B medical models and approaches the 72B general model.}
\label{tab:performance_ceiling}
\end{table*}

\section{Quantitative Assessment of Hallucinations Beyond LLM-as-a-Judge}
\label{ap:hallucination}

To provide a more comprehensive assessment and move beyond relying solely on LLM-judges, we conducted an additional rigorous hallucination analysis using two complementary, established benchmarks:
\begin{itemize}
    \item \textbf{TruthfulQA (Medical subset) \cite{lin2022truthfulqa}:} A gold standard for evaluating general truthfulness and the model's susceptibility to common misconceptions.
    \item \textbf{MedHALT (Fake Question subset) \cite{pal2023med}:} A specialized medical benchmark designed specifically to test ``confabulation'', the tendency of models to generate realistic-looking answers for non-existent or fabricated medical queries.
\end{itemize}

We report the Hallucination Rate (where lower is better, denoted by $\downarrow$) in Table \ref{tab:hallucination_results}. SAGE consistently maintains the lowest hallucination rate across both benchmarks. This confirms that our method effectively calibrates the model's factual boundaries by grounding generation in MeSH taxonomies, rather than relying on unconstrained self-generation.

\begin{table}[]
\centering
\begin{tabular}{llcc}
\toprule
\textbf{Strategy} & \textbf{TruthfulQA $\downarrow$} & \textbf{MedHALT $\downarrow$} \\
\midrule
Wrap              & 32.66 & 90.84 \\
CoT-SI            & 22.61 & 91.59 \\
\textbf{SAGE}     & \textbf{20.60} & \textbf{89.28} \\
\bottomrule
\end{tabular}
\caption{Hallucination rates on TruthfulQA (Medical subset) and MedHALT (Fake Question subset). Lower values indicate better factuality. All trainee models are based on Qwen3-4B-Base. ``CoT-SI'' stands for CoT-Self-Instruct. Results are reported in percentages (\%).}
\label{tab:hallucination_results}
\end{table}

\section{In-depth Comparison with Document-Based Methods}
\label{ap:corpus_ablation}

Our problem setting is specifically designed to simulate a practical, resource-constrained scenario (e.g., a small-scale specialized clinic), where massive, clean medical corpora are often inaccessible due to privacy or computing limitations. In this context, the MeSH taxonomy serves as a foundational medical dictionary, while the 1K seed Q\&A samples represent years of accumulated patient case records. Our initial baseline comparisons utilized MeSH Scope Notes as the primary corpus for document-based methods (e.g., Genie) specifically to ensure strict fairness in knowledge sources, preventing any unintended advantage from external data.

To comprehensively assess whether the performance gap between SAGE and document-based methods stems from the brevity of scope notes rather than the synthesis paradigm itself, we conducted two rigorous ablation experiments:
\begin{enumerate}
    \item \textbf{Data Augmentation:} We augmented the document-based baseline with the identical 1K seed Q\&A cases used by SAGE.
    \item \textbf{Corpus Substitution:} We replaced the scope notes entirely with a traditional medical corpus (PubMed abstracts), extracting an equivalent volume of documents to strictly match SAGE's 3.51M token budget.
\end{enumerate}

As shown in Table \ref{tab:corpus_ablation}, simply scaling the token budget or switching to traditional medical documents yields only marginal improvements for the document-based baseline, which remains significantly inferior to SAGE. This demonstrates that SAGE's superiority does not stem from simply exposing the model to more text. Instead, its effectiveness lies in its unique ability to distill static, dictionary-style facts into dense, high-quality clinical reasoning instructions, which is far more effective for fine-tuning small LLMs in data-scarce environments.

\begin{table*}[]
\centering
\resizebox{\textwidth}{!}{
\begin{tabular}{llccccccc}
\toprule
\textbf{Strategy} & \textbf{Data Source} & \textbf{Tokens} & \textbf{MedMCQA} & \textbf{MedQA} & \textbf{PubMedQA} & \textbf{MMLU-Pro} & \textbf{GPQA} & \textbf{AVG} \\
\midrule
Genie         & Scope Notes           & 1.69M          & 49.32 & 51.77 & 63.90 & 48.99 & 35.90 & 49.98 \\
Genie         & Notes + Distilled Q\&A& 3.51M          & 46.86 & 55.15 & 75.30 & 52.31 & 35.13 & 52.95 \\
Genie         & PubMed Abstracts      & 3.51M          & 52.38 & 48.63 & 61.00 & 51.21 & 40.00 & 50.64 \\
\midrule
\textbf{SAGE} & \textbf{Notes + Distilled Q\&A} & \textbf{3.51M} & \textbf{64.98} & \textbf{70.07} & \textbf{80.80} & \textbf{63.19} & \textbf{61.54} & \textbf{68.12} \\
\bottomrule
\end{tabular}
}
\caption{Ablation study on data sources and token budgets. All trainee models are Qwen3-4B-Base. MMLU-Pro and GPQA refer to their respective medical subsets. Results are reported in percentages (\%).}
\label{tab:corpus_ablation}
\end{table*}

\section{Robustness and Evaluation of the Mapping Function}
\label{ap:mapping_evaluation}

The mapping function $\phi$ is fundamentally grounded in string matching, which guarantees its lightweight nature and extremely low computational overhead. However, to ensure high clinical specificity and robustness against extraction errors, $\phi$ extends significantly beyond naive string matching by employing a \textit{synonym-expanded matching and structural filtering} pipeline. Specifically, we incorporate the comprehensive MeSH synonym dictionary to capture semantic variations during the matching process. Furthermore, to eliminate shallow, noisy vocabulary (e.g., broad terms like ``Disease''), we strictly filter out terms with a MeSH tree depth $< 4$ and restrict anchors exclusively to core medical categories (e.g., Anatomy, Diseases). This design ensures that all extracted semantic anchors maintain high clinical specificity without sacrificing processing speed.

\paragraph{Quantitative Error Analysis}

To provide a quantitative evaluation of the concept extraction, we analyzed 1,000 seed samples using the official MeSH on Demand (MoD) tool, a sophisticated online engine utilizing MetaMap and PubMed k-NN mapping which can be used as the pseudo-gold standard. 

As shown in Table \ref{tab:mapping_precision_recall}, while naive string matching achieves a higher recall, it suffers from remarkably low precision (36.52\%). This indicates that while it successfully extracts many terms, the anchor set is heavily flooded with clinically uninformative vocabulary. SAGE intentionally trades off the recall of these broad terms (12.21\%) to maximize the clinical density and precision (55.68\%) of the extracted anchors. In the context of data synthesis, feeding small LLMs with a few highly precise, deep semantic anchors is vastly superior to flooding them with numerous shallow, noisy concepts. Thus, the high precision of our $\phi$ function safeguards the structural integrity and clinical density of the generation trajectories.

\begin{table}[]
\centering
\begin{tabular}{lcc}
\toprule
\textbf{Extraction Method} & \textbf{Precision} & \textbf{Recall} \\
\midrule
Naive String Matching & 36.52 & 55.30 \\
SAGE Mapping (Ours) & 55.68 & 12.21 \\
\bottomrule
\end{tabular}
\caption{Quantitative evaluation of extraction methods against the MeSH on Demand pseudo-gold standard. Results are reported in percentages (\%).}
\label{tab:mapping_precision_recall}
\end{table}

\paragraph{Downstream Performance and Computational Overhead}

To definitively validate the robustness of $\phi$, we compared the downstream QA performance, privacy constraints, and computational overhead of these extraction strategies. All experiments utilized Qwen3-4B-Base as the trainee model.

As Table \ref{tab:mapping_downstream} illustrates, naive string matching performs poorly (64.02\% average) due to the interference of noisy anchors. Conversely, while the official MoD tool yields strong downstream results, it requires continuous external network requests, violating the strict data privacy setting of practical clinical environments, and incurs an immense time overhead ($\sim$40$\times$ slower), making it unscalable for large-scale data synthesis. Our SAGE mapping achieves virtually identical downstream performance (68.12\% vs. 68.14\%) compared to the heavy online expert tool, but executes securely offline at a fraction of the computational cost. This confirms that our framework is highly robust to extraction noise and optimally designed for resource-constrained, privacy-sensitive scenarios.

\begin{table*}[!h]
\centering
\resizebox{\textwidth}{!}{
\begin{tabular}{llccccccc}
\toprule
\textbf{Extraction Method} & \textbf{Avg Time / Sample} & \textbf{Privacy} & \textbf{MedMCQA} & \textbf{MedQA} & \textbf{PubMedQA} & \textbf{MMLU-Pro} & \textbf{GPQA} & \textbf{AVG} \\
\midrule
Naive Matching  & 1.15s  & Offline    & 63.81 & 69.60 & 70.40 & 61.95 & 54.36 & 64.02 \\
MeSH on Demand  & 48.24s & Online API & 66.08 & 67.16 & 83.20 & 62.74 & 61.54 & 68.14 \\
\midrule
\textbf{SAGE Mapping} & \textbf{1.31s} & \textbf{Offline} & \textbf{64.98} & \textbf{70.07} & \textbf{80.80} & \textbf{63.19} & \textbf{61.54} & \textbf{68.12} \\
\bottomrule
\end{tabular}
}
\caption{Comparison of extraction methods regarding downstream QA performance, computational overhead, and privacy. MMLU-Pro and GPQA refer to their respective medical subsets. Downstream metrics are reported in percentages (\%).}
\label{tab:mapping_downstream}
\end{table*}

\section{Ablation on Factual Grounding and Filtering Strategies}
\label{ap:filtering}

While self-consistency voting is a widely adopted proxy for uncertainty estimation in resource-constrained settings (e.g., in CoT-Self-Instruct), it serves primarily as a confidence filter to prune unstable reasoning paths and does not necessarily guarantee factual accuracy. If a model exhibits systematic knowledge gaps, repeated reasoning paths may converge on the same incorrect answer, potentially allowing consistently erroneous samples to pass the filtering process.

To prevent these systematic knowledge gaps, SAGE does not rely solely on post-hoc filtering. Through Atomic Generation, structured factual information (MeSH definitions) is explicitly incorporated into the generation trajectory. This ensures the synthesis is grounded in authoritative knowledge from the start, mitigating the risk of the model consistently repeating its internal flawed guesses.

To validate this design choice, we tested an alternative approach using contradiction filtering, where MeSH Scope Notes were used as a reference for the small model to post-verify its outputs. As shown in Table \ref{tab:filtering_ablation}, this approach yielded inferior results. In resource-constrained settings, small models lack the discriminative capacity to accurately detect subtle semantic contradictions. Consequently, it is more effective to embed factual knowledge directly into the training data generation phase and use self-consistency to reinforce the model's confidence.

\begin{table*}[!h]
\centering
\resizebox{\textwidth}{!}{
\begin{tabular}{lcccccc}
\toprule
\textbf{Filtering Strategy} & \textbf{MedMCQA} & \textbf{MedQA} & \textbf{PubMedQA} & \textbf{MMLU-Pro} & \textbf{GPQA} & \textbf{AVG} \\
\midrule
Contradiction Filtering & 64.67 & 68.03 & 70.80 & 60.72 & 58.46 & 64.54 \\
\textbf{SAGE (Consistency + Atomic Gen.)} & \textbf{64.98} & \textbf{70.07} & \textbf{80.80} & \textbf{63.19} & \textbf{61.54} & \textbf{68.12} \\
\bottomrule
\end{tabular}
}
\caption{Performance comparison between post-hoc contradiction filtering and the SAGE approach. MMLU-Pro and GPQA refer to their respective medical subsets. Results are reported in percentages (\%).}
\label{tab:filtering_ablation}
\end{table*}

\section{Qualitative Analysis of Baseline Methods}
\label{ap:baseline_case}
In this section, we present a detailed case study illustrating the limitations of existing synthesis paradigms discussed in Section~\ref{sec:qual_case}.
\begin{figure*}[!h]
    \centering
    \includegraphics[width=0.98\linewidth]{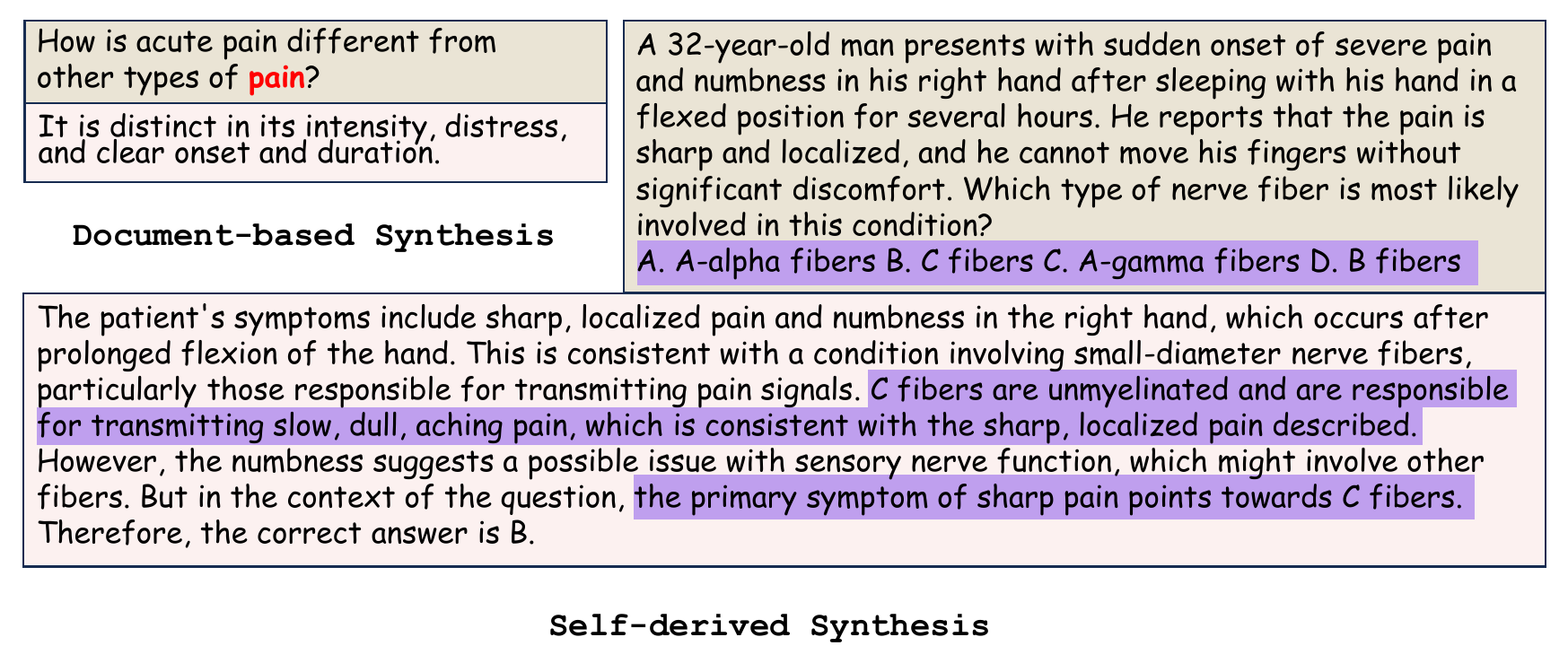}
    \caption{A detailed case on self-derived and document-based synthesis.}
    \label{fig:two_baselines_case}
\end{figure*}
As illustrated in Figure~\ref{fig:two_baselines_case}, the Wrap method, representing document-based synthesis, generates a simplistic question about distinguishing ``acute pain'' from other types of pain. The corresponding response, while factually inoffensive, is superficial and of limited educational utility. On the other hand, CoT-Self-Instruct, representing self-derived synthesis, produces a question about the functions and clinical localization of different nerve fiber types. However, due to insufficient factual grounding, it fails to include A-delta fibers as a valid option within the appropriate clinical context, and incorrectly attributes sharp pain to C fibers. This constitutes a serious hallucination that contradicts fundamental neurophysiological consensus.

\section{Evaluation on Extended Benchmarks}
\label{ap:extended_benchmark}
We evaluated all baseline methods on the extended benchmarks and compiled their performance. The results, presented in Figure~\ref{tab:extended}, demonstrate that our approach consistently achieves leading performance.

\begin{table*}[!h]
\centering
\label{tab:appendix_subset_results}
\resizebox{\textwidth}{!}{%
\begin{tabular}{l l c | c c | c c c c c}
\toprule
\textbf{Model} & \textbf{Method/Dataset} & \textbf{Scale} & \textbf{AVG-5} & \textbf{AVG-10} & \textbf{Lancet} & \textbf{MedBul\_op4} & \textbf{MedBul\_op5} & \textbf{MedXpert} & \textbf{NEJM} \\
\midrule
Qwen3-4B-Base & Few-shot Prompting & NA & 26.15 & 24.88 & 30.58 & 29.55 & 24.03 & 11.18 & 22.72 \\
Qwen3-4B NoThinking & NA & NA & 62.54 & 54.15 & 59.47 & 50.32 & 48.05 & 13.04 & 57.88 \\
Qwen3-4B Thinking & NA & NA & \underline{66.18} & \underline{57.98} & 62.86 & 59.09 & 50.97 & 13.73 & 62.19 \\
\midrule
\multicolumn{10}{l}{\textit{Open Source Datasets}} \\
Qwen3-4B-Base & m1k & 1k & 63.40 & 56.98 & 61.65 & 59.74 & 54.87 & 17.05 & 59.54 \\
Qwen3-4B-Base & m23k & 23k & \underline{67.39} & \underline{61.44} & 65.05 & 67.53 & 61.04 & 20.50 & 63.35 \\
Qwen3-4B-Base & HuatuoGPT-o1-SFT & 20k & 52.80 & 48.22 & 56.31 & 51.62 & 43.51 & 15.73 & 51.08 \\
Qwen3-4B-Base & MedReason & 33k & 58.60 & 51.76 & 51.94 & 54.22 & 49.35 & 16.84 & 52.24 \\
\midrule
\multicolumn{10}{l}{\textit{Self-derived Method}} \\
Qwen3-4B-Base & self-instruct & 10k & 53.66 & 47.26 & 50.73 & 47.08 & 41.23 & 13.18 & 52.07 \\
Qwen3-4B-Base & scalequest & 10k & 62.91 & 54.17 & 58.25 & 51.30 & 47.40 & 13.66 & 56.55 \\
Qwen3-4B-Base & KPDDS & 10k & 58.99 & 50.26 & 53.88 & 44.16 & 41.56 & 13.87 & 54.23 \\
Qwen3-4B-Base & MathScale & 10k & 57.85 & 50.37 & 59.22 & 47.73 & 39.29 & 14.63 & 53.57 \\
Qwen3-4B-Base & Cot-Self-Instruct & 10k & \underline{63.71} & \underline{56.80} & 64.32 & 57.14 & 48.70 & 19.12 & 60.20 \\
\midrule
\multicolumn{10}{l}{\textit{Document-based Method}} \\
Qwen3-4B-Base & Easy-Dataset & 28k & 49.11 & 41.40 & 44.17 & 37.34 & 34.09 & 11.32 & 41.46 \\
Qwen3-4B-Base & Genie & 30k & 49.98 & 44.87 & 54.85 & 44.16 & 37.99 & 14.22 & 47.60 \\
Qwen3-4B-Base & Self-QA & 230k & 46.34 & 42.62 & 50.73 & 42.21 & 39.29 & 13.66 & 48.59 \\
Qwen3-4B-Base & Wrap & 98K & \underline{50.20} & \underline{45.43} & 53.40 & 44.48 & 42.21 & 15.11 & 48.09 \\
\midrule
\multicolumn{10}{l}{\textit{Ours}} \\
Qwen3-4B-Base & w/o Iteration & 10k & 66.45 & 58.51 & 63.83 & 58.12 & 50.32 & 21.39 & 59.20 \\
Qwen3-4B-Base & w/o Parallel & 10k & 66.81 & 58.33 & 62.38 & 57.14 & 48.38 & 23.12 & 58.21 \\
Qwen3-4B-Base & w/o Atomic gen. & 10k & 66.21 & 58.17 & 62.14 & 57.47 & 51.30 & 22.15 & 57.55 \\
Qwen3-4B-Base & w/o Associative gen. & 10k & 66.13 & 59.62 & 68.20 & 59.09 & 52.27 & 23.60 & 62.35 \\
Qwen3-4B-Base & \textbf{Ours(Full)} & 10k & \underline{\textbf{68.12}} & \underline{\textbf{59.79}} & 66.02 & 59.42 & 46.43 & 22.71 & 62.69 \\
\bottomrule
\end{tabular}%
}
\caption{Performance comparison on additional datasets. Since the first five medical datasets are detailed in the main text, this table focuses on the remaining datasets and the overall AVG-10 scores (\%).}
\label{tab:extended}
\end{table*}

\section{Token Efficiency Analysis}
\label{ap:efficiency}

\begin{figure}
    \centering
    \includegraphics[width=0.98\linewidth]{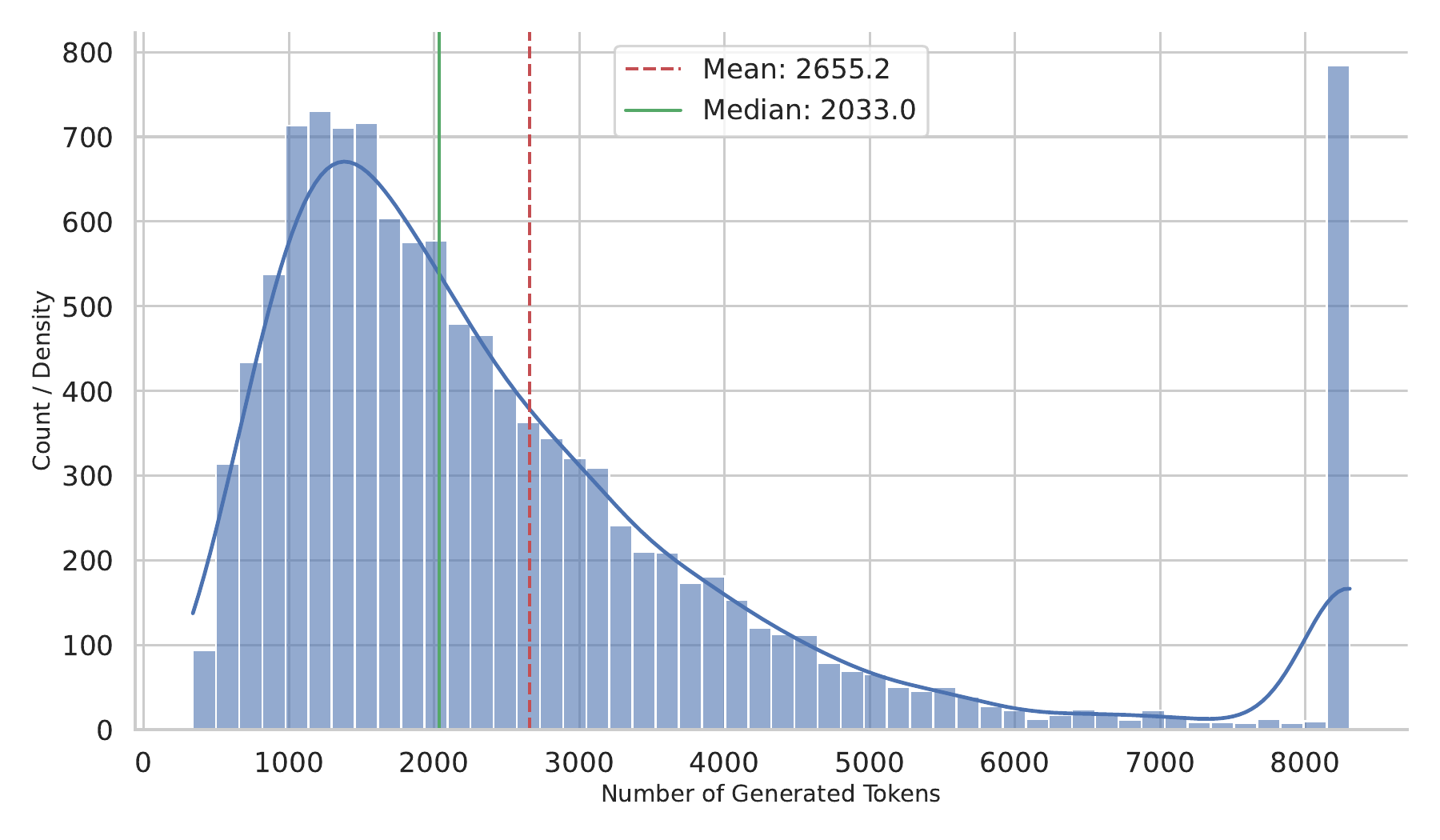}
    \caption{Distribution chart of token lengths in the test set on Qwen3-4B with thinking mode.}
    \label{fig:token_4b_think}
\end{figure}

\begin{figure}
    \centering
    \includegraphics[width=0.98\linewidth]{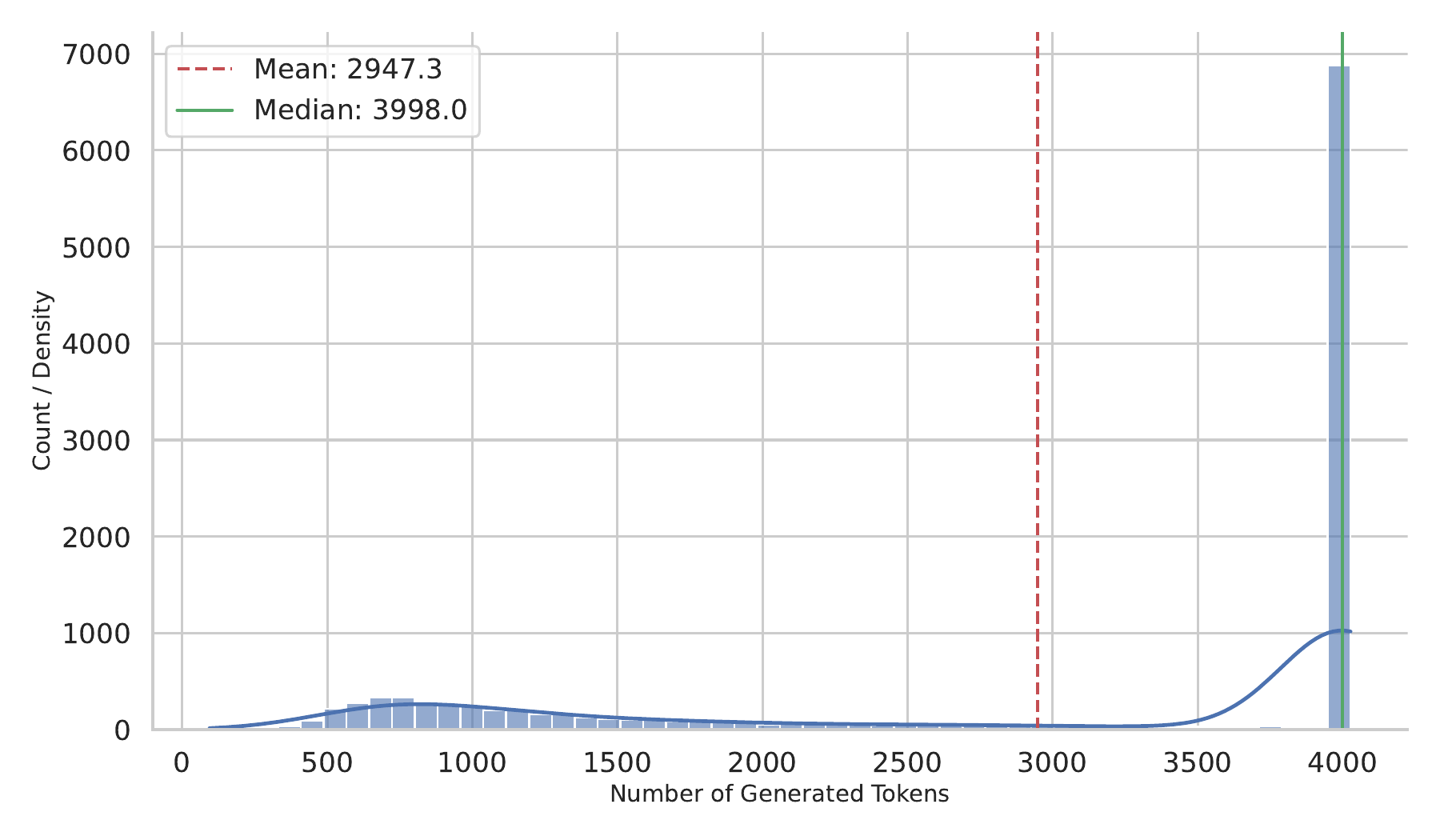}
    \caption{Distribution chart of token lengths in the test set after training with m23k dataset on Qwen3-4B-Base.}
    \label{fig:token_m23k}
\end{figure}

\begin{figure}
    \centering
    \includegraphics[width=0.98\linewidth]{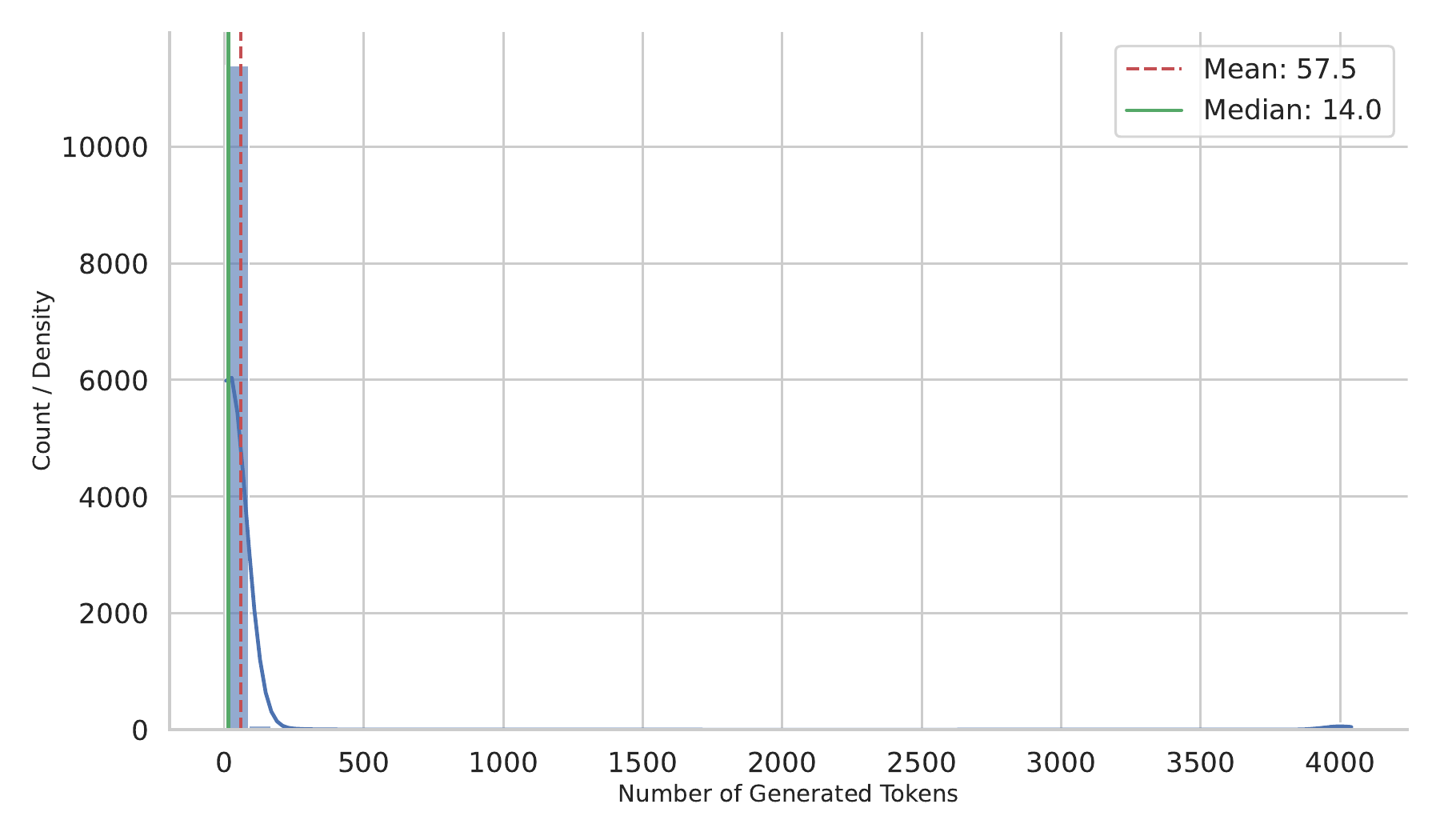}
    \caption{Distribution chart of token lengths in the test set after training the data synthesized by the Wrap method on Qwen3-4B-Base.}
    \label{fig:token_wrap}
\end{figure}

\begin{figure}
    \centering
    \includegraphics[width=0.98\linewidth]{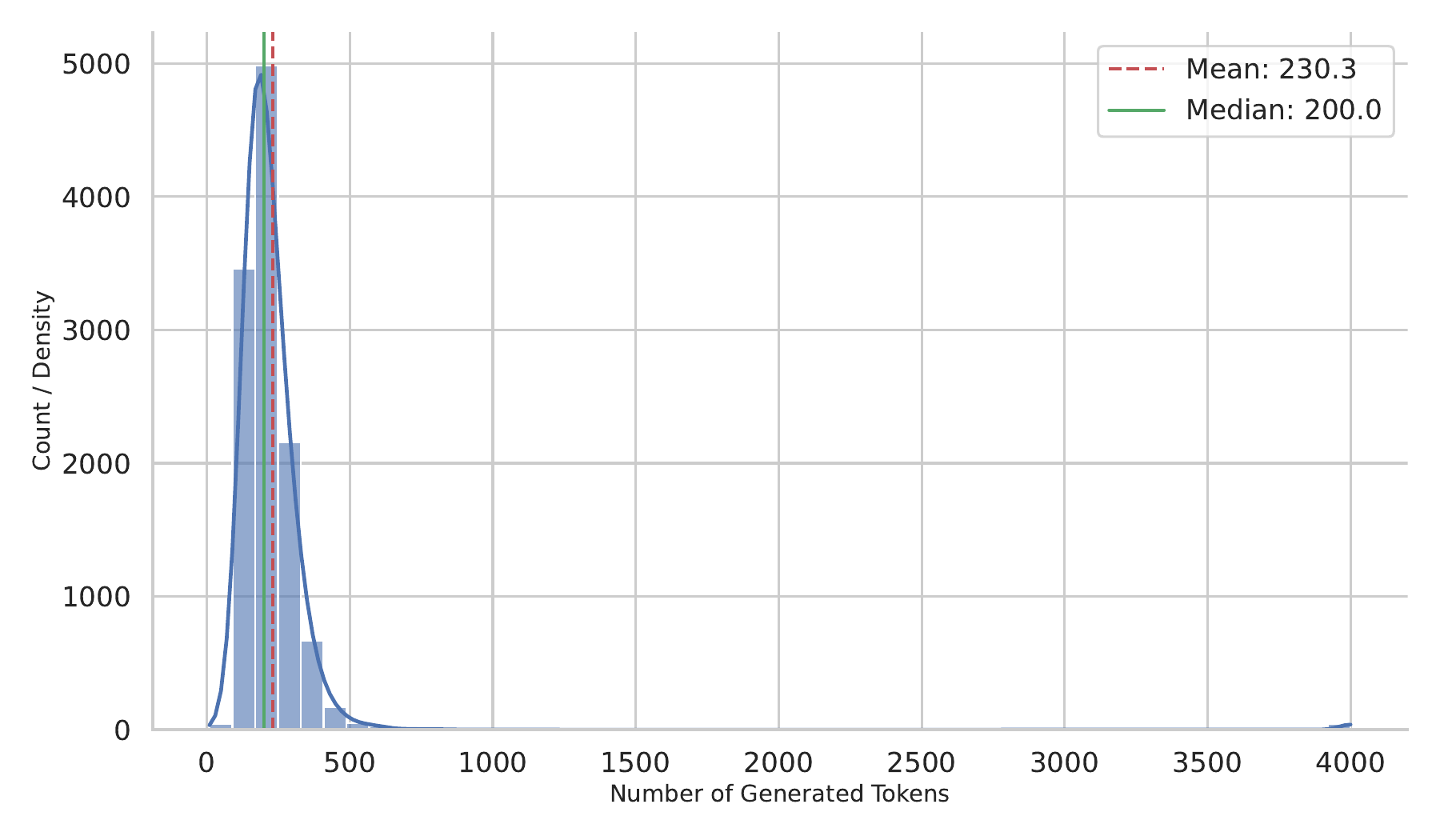}
    \caption{Distribution chart of token lengths in the test set after training the data synthesized by our method on Qwen3-4B-Base.}
    \label{fig:token_msi}
\end{figure}

To demonstrate the token efficiency of our method, we compared the token lengths of inference responses across multiple baselines on ten benchmark test sets (results shown in Figures~\ref{fig:token_4b_think}, \ref{fig:token_m23k}, \ref{fig:token_wrap}, and \ref{fig:token_msi}). Our method uses the non-thinking mode of Qwen3-4B for data synthesis, mainly considering the efficiency and practicality of offline data generation under resource-constrained settings. We further compare thinking and non-thinking modes for data synthesis using both SAGE and CoT-Self-Instruct, with each setting repeated over three runs. As shown in Table~\ref{tab:thinking_vs_nonthinking}, using thinking mode during data synthesis does not bring consistent improvements. As illustrated, our approach not only achieves state-of-the-art performance as shown in Section~\ref{sec:main_results}, but also maintains significantly lower token lengths compared to all baseline methods. Specifically, our method yields an average token length of only $230.3$, which is notably lower than the $2655.2$ tokens of Qwen3-4B's thinking mode and the $2947.3$ tokens of the M23K dataset distilled from R1-style responses. Furthermore, the Wrap method exhibits excessively short token lengths, averaging only $57.5$, which aligns with the analysis presented in Appendix~\ref{ap:baseline_case}. Therefore, our method achieves a favorable balance between downstream effectiveness and inference-time token efficiency.

\begin{table*}[t]
\centering
\small
\resizebox{\linewidth}{!}{
\begin{tabular}{lcccccc}
\toprule
\textbf{Method} & \textbf{AVG-5} & \textbf{MedMCQA} & \textbf{MedQA} & \textbf{PubMedQA} & \textbf{MMLU-Pro (med)} & \textbf{GPQA (med)} \\
\midrule
SAGE (thinking) 
& $67.85 \pm 0.26$ 
& $65.11 \pm 0.89$ 
& $67.61 \pm 0.39$ 
& $82.27 \pm 0.81$ 
& $62.48 \pm 0.75$ 
& $61.80 \pm 1.18$ \\
SAGE (non-thinking) 
& $68.05 \pm 0.22$ 
& $65.26 \pm 0.98$ 
& $68.24 \pm 1.85$ 
& $82.03 \pm 0.98$ 
& $62.21 \pm 1.02$ 
& $62.50 \pm 2.31$ \\
\midrule
CoT-Self-Instruct (thinking) 
& $63.57 \pm 0.50$ 
& $62.34 \pm 0.13$ 
& $68.34 \pm 0.95$ 
& $69.63 \pm 0.23$ 
& $60.98 \pm 0.55$ 
& $56.58 \pm 0.97$ \\
CoT-Self-Instruct (non-thinking) 
& $63.75 \pm 0.10$ 
& $61.47 \pm 0.92$ 
& $67.11 \pm 0.94$ 
& $74.30 \pm 4.07$ 
& $62.26 \pm 0.44$ 
& $53.59 \pm 3.20$ \\
\bottomrule
\end{tabular}
}
\caption{
Comparison between thinking and non-thinking modes for data synthesis. 
All results are reported as mean $\pm$ standard deviation over three runs. 
AVG-5 denotes the average score over the five medical benchmarks (\%).
}
\label{tab:thinking_vs_nonthinking}
\end{table*}

\section{Stability Across Independent Runs}
\label{app:stability}

To further examine the stability of SAGE, we conduct three independent runs for SAGE and two primary synthetic-data baselines, CSI (CoT-Self-Instruct) and WRAP, using the same experimental setting. Table~\ref{tab:stability_runs} reports the mean accuracy and standard deviation across five medical QA benchmarks.

As shown in the table, SAGE consistently achieves the best average performance across the evaluated benchmarks across runs. This suggests that the observed improvements are not driven by a single favorable run, but reflect stable gains from the proposed grounded synthesis framework.

\begin{table*}[t]
\centering
\small
\begin{tabular}{lcccccc}
\toprule
\textbf{Method} & \textbf{AVG-5} & \textbf{MedMCQA} & \textbf{MedQA} & \textbf{PubMedQA} & \textbf{MMLU-Pro (med)} & \textbf{GPQA (med)} \\
\midrule
SAGE & $68.05 \pm 0.22$ & $65.26 \pm 0.98$ & $68.24 \pm 1.85$ & $82.03 \pm 0.98$ & $62.21 \pm 1.02$ & $62.50 \pm 2.31$ \\
CSI & $63.75 \pm 0.10$ & $61.47 \pm 0.92$ & $67.11 \pm 0.94$ & $74.30 \pm 4.07$ & $62.26 \pm 0.44$ & $53.59 \pm 3.20$ \\
WRAP & $49.18 \pm 0.95$ & $47.81 \pm 0.32$ & $47.03 \pm 0.79$ & $66.37 \pm 3.10$ & $46.40 \pm 0.69$ & $38.29 \pm 1.41$ \\
\bottomrule
\end{tabular}
\caption{Mean accuracy and standard deviation over three independent runs. SAGE shows consistently strong performance with low variance across five medical QA benchmarks (\%).}
\label{tab:stability_runs}
\end{table*}

\section{Comparison with Fine-Tuning on Original Training Sets}
\label{app:original_training_sets}

We also compare SAGE with supervised fine-tuning on the original training sets of datasets that provide dedicated training splits. Specifically, we fine-tune the same base LLM on the training sets of MedMCQA, MedQA, and PubMedQA, which contain approximately 10k, 100k, and 200k samples, respectively. 

Table~\ref{tab:original_training_sets} reports the mean accuracy and standard deviation across multiple runs. Despite using existing human-curated training data, fine-tuning on the original dataset training splits yields substantially lower performance than SAGE. One possible reason is that these datasets were not primarily designed for instruction tuning of LLMs: their target outputs are often short answer-only labels, providing limited supervision for learning reasoning-oriented response patterns. In contrast, SAGE constructs explanation-rich instruction data grounded in semantic anchors, which provides stronger supervision for medical question answering.

\begin{table}[t]
\centering
\small
\begin{tabular}{lcc}
\toprule
\textbf{Dataset} & \textbf{Train-Set} & \textbf{SAGE} \\
\midrule
MedMCQA & $57.42 \pm 0.76$ & $65.26 \pm 0.98$ \\
MedQA & $56.79 \pm 1.85$ & $68.24 \pm 1.85$ \\
PubMedQA & $50.60 \pm 1.12$ & $82.03 \pm 0.98$ \\
\bottomrule
\end{tabular}
\caption{Comparison between fine-tuning on original dataset training splits and fine-tuning on SAGE-generated data. Results are reported as mean accuracy and standard deviation across multiple runs (\%).}
\label{tab:original_training_sets}
\end{table}


\section{Prompt Templates}
\label{ap:prompt}
To diversify our iterative data generation process, we employed two distinct question types: atomic generation and associative generation. The specific prompt templates used are provided in Figure~\ref{fig:prompt_as} and Figure~\ref{fig:prompt_at}, respectively.

\begin{figure}[]
    \centering
    \begin{tcolorbox}[
        colback=gray!5,           
        colframe=gray!60,         
        title=\textbf{Prompt Template for Associative Generation} , 
        arc=1mm,                  
        boxrule=0.8pt,            
        width=\linewidth,         
        fontupper=\scriptsize\ttfamily, 
        left=2pt, right=2pt,      
        top=2pt, bottom=2pt
    ]
        You are a medical expert creating high-quality multiple-choice questions for medical education.

        \vspace{1em}
        Few-shot examples:\\
        Example 1 (Concepts: \textit{\{terms\_str\}}):\\
        \textit{\{q\_text\}}

        \vspace{0.5em}
        Example 2 (Concepts: \textit{\{terms\_str\}}):\\
        \textit{\{q\_text\}}

        \vspace{1em}
        Task: Generate a NEW medical multiple-choice question that meaningfully incorporates the concepts from these MeSH terms: \textit{\{target\_terms\}}

        \vspace{1em}
        Requirements:
        \begin{enumerate}
            \item[1.] Create a medically relevant, challenging question
            \item[2.] Include 3-6 plausible options labeled A, B, C, D, ...
            \item[3.] Only ONE option should be correct
            \item[4.] The question should require medical reasoning, not just recall
            \item[5.] Do NOT copy from the examples
        \end{enumerate}

        \vspace{0.5em}
        Output format STRICTLY:\\
        {[}Question Begin{]}\\
        \{\{your generated question with options\}\}\\
        {[}Question End{]}\\
        {[}Reasoning Begin{]}\\
        \{\{step-by-step medical reasoning\}\}\\
        {[}Reasoning End{]}\\
        {[}Answer Begin{]}\textbackslash boxed\{\{correct option letter\}\}{[}Answer End{]}
        
    \end{tcolorbox}
    \caption{The prompt template used for associative generation. Placeholders denote: \textit{\{terms\_str\}} (concepts associated with the few-shot examples), \textit{\{q\_text\}} (the content of the few-shot example questions), and \textit{\{target\_terms\}} (the set of MeSH terms to be incorporated into the new question).}
    \label{fig:prompt_as}
\end{figure}

\begin{figure}[]
    \centering
    \begin{tcolorbox}[
        colback=gray!5,           
        colframe=gray!60,         
        title=\textbf{Prompt Template for Atomic Generation}, 
        arc=1mm,                  
        boxrule=0.8pt,            
        width=\linewidth,         
        fontupper=\scriptsize\ttfamily, 
        left=2pt, right=2pt,      
        top=2pt, bottom=2pt
    ]
        You are a medical expert.
        
        \vspace{0.3em}
        Target Concept: \textit{\{term\}}\\
        Definition: \textit{\{scope\_note\}}\\
        Relevant Aspects: \textit{\{qual\_str\}}

        \vspace{0.5em}
        Task: Create a detailed multiple-choice question specifically focused on the \textit{\{term\}}.

        \vspace{0.5em}
        Output Format:\\
        {[}Question Begin{]}...{[}Question End{]}\\
        {[}Reasoning Begin{]}...{[}Reasoning End{]}\\
        {[}Answer Begin{]}\textbackslash boxed\{\{A\}\}{[}Answer End{]}
        
    \end{tcolorbox}
    \caption{The prompt template used for atomic generation. Placeholders denote: \textit{\{term\}} (the target medical concept), \textit{\{scope\_note\}} (the official definition/explanation of the term), and \textit{\{qual\_str\}} (relevant sub-aspects or qualifiers associated with the term).}
    \label{fig:prompt_at}
\end{figure}

\end{document}